\documentclass[runningheads]{llncs}
\usepackage{graphicx}
\usepackage{multirow}
\usepackage{comment}
\usepackage{amsmath,amssymb} % define this before the line numbering.
\usepackage{color}
\usepackage{adjustbox}
\usepackage{pifont}
\usepackage{tabularx}
\usepackage{cite}
\usepackage{hyperref}
\hypersetup{colorlinks}
\newcommand{\cmark}{\text{\ding{51}}}

\newcommand{\widthscalefive}{0.145}
\usepackage[width=122mm,left=12mm,paperwidth=146mm,height=193mm,top=12mm,paperheight=217mm]{geometry}

\begin{document}
% \renewcommand\thelinenumber{\color[rgb]{0.2,0.5,0.8}\normalfont\sffamily\scriptsize\arabic{linenumber}\color[rgb]{0,0,0}}
% \renewcommand\makeLineNumber {\hss\thelinenumber\ \hspace{6mm} \rlap{\hskip\textwidth\ \hspace{6.5mm}\thelinenumber}}
% \linenumbers
\pagestyle{headings}
\mainmatter

\title{Pyramid Attention Networks \\ for Image Restoration} % Replace with your title

% INITIAL SUBMISSION 
%\begin{comment}
\author{\scriptsize Yiqun Mei$^{1}$, Yuchen Fan$^{1}$, Yulun Zhang$^{2}$, Jiahui Yu$^{3}$, Yuqian Zhou$^{1}$, Ding Liu$^{4}$, \\ Yun Fu$^{2}$,
Thomas S. Huang$^{1}$, Humphrey Shi$^{5,1}$\\
}
\institute{\scriptsize $^1$IFP Group, UIUC, $^2$SmileLab, Northeastern University, $^3$Google Brain,\\ $^4$ByteDance AI Lab, $^5$ University of Oregon}
%\end{comment}
%******************

% CAMERA READY SUBMISSION
\begin{comment}
\titlerunning{Abbreviated paper title}
% If the paper title is too long for the running head, you can set
% an abbreviated paper title here
%
\author{First Author\inst{1}\orcidID{0000-1111-2222-3333} \and
Second Author\inst{2,3}\orcidID{1111-2222-3333-4444} \and
Third Author\inst{3}\orcidID{2222--3333-4444-5555}}
%
\authorrunning{F. Author et al.}
% First names are abbreviated in the running head.
% If there are more than two authors, 'et al.' is used.
%
\institute{Princeton University, Princeton NJ 08544, USA \and
Springer Heidelberg, Tiergartenstr. 17, 69121 Heidelberg, Germany
\email{lncs@springer.com}\\
\url{http://www.springer.com/gp/computer-science/lncs} \and
ABC Institute, Rupert-Karls-University Heidelberg, Heidelberg, Germany\\
\email{\{abc,lncs\}@uni-heidelberg.de}}
\end{comment}
%******************
\maketitle
%#### Abstract ####
\begin{abstract}
% Self-similarity that small patterns tend to occur at different locations and scales has been widely explored by classic image restoration algorithms. However, recent advanced deep convolution neural networks relying on convolution operations or non-local attentions only process information at a same scale. To solve this problem, we present a novel pyramid attention for image restoration, which captures long-range feature correspondences from a multi-scale feature pyramid. Inspired by the fact that corruption effects, such as noise or compression artifacts, drop drastically at coarser image scales, the attention is designed to borrow clean signals from their ``cleane'' correspondences at coarser levels. The proposed pyramid attention is a generic building block that can be flexibly integrated into many architectures. We conduct extensive experiments on image denoising, demosaicing ,compression artifact reduction and image super resolution. Even without any bells and whistles, our pyramid attention with a simple ResNet backbone can produce state-of-the-art results with superior accuracy and visual quality. 
Self-similarity refers to the image prior widely used in image restoration algorithms that small but similar patterns tend to occur at different locations and scales. However, recent advanced deep convolutional neural network based methods for image restoration do not take full advantage of self-similarities by relying on self-attention neural modules that only process information at the same scale. 
To solve this problem, we present \textbf{a novel Pyramid Attention module} for image restoration, which captures long-range feature correspondences from a multi-scale feature pyramid. Inspired by the fact that corruptions, such as noise or compression artifacts, drop drastically at coarser image scales, our attention module is designed to be able to \textit{borrow} clean signals from their ``clean'' correspondences at the coarser levels. The proposed pyramid attention module is a generic building block that can be flexibly integrated into \textbf{various neural architectures}. Its effectiveness is validated through extensive experiments \textbf{on multiple image restoration tasks}: image denoising, demosaicing, compression artifact reduction, and super resolution. Without any bells and whistles, our \textbf{PANet} (pyramid attention module with simple network backbones) can produce state-of-the-art results with superior accuracy and visual quality. Our code will be available at \href{https://github.com/SHI-Labs/Pyramid-Attention-Networks}{https://github.com/SHI-Labs/Pyramid-Attention-Networks}
\keywords{Image Restoration, Pyramid Attention, Image Denoising, Demosaicing, Compression Artifact Reduction, Super Resolution}
\end{abstract}

%#### Introduction ####
\section{Introduction} \label{sec:1}
% Image restoration aims to recover a high-quality image from a contaminated counterpart, which is an ill-posed problem due to the irreversible degradation processes. Depending on the type of corruptions, such problem can be generally classified into several categories: image denoising, demosaicing and compression artifacts reduction. A variety of classic approaches has been proposed to solve this problem. 
Image restoration algorithms aim to recover a high-quality image from the contaminated counterpart, and is viewed as an ill-posed problem due to the irreversible degradation processes. They have many applications depending on the type of corruptions, for example, image denoising \cite{zhang2017beyond,zhang2019residual,liu2018non}, demosaicing \cite{zhang2017learning,zhang2019residual}, compression artifacts reduction \cite{dong2015compression,chen2017trainable,zhang2017beyond}, super resolution \cite{kim2016accurate,lai2017deep,tai2017memnet} and many others \cite{li2017aod,he2010single,chen2018learning}. To restore missing information in the contaminated image, a variety of approaches based on leveraging image priors have been proposed \cite{buades2005non,zontak2013separating,roth2005fields,zoran2011learning}.

%\textbf{Non-local prior in classical methods is useful. Cross-scale non-local are used for SR, Denoise,etc}
% One of the most representative ways is to leverage image priors.
Among these approaches, the prior of self-similarity in an image is widely explored and proven to be important. For example, non-local mean filtering \cite{buades2005non} uses self-similarity prior to reduce corruptions, which averages similar patches within the image. This notion of non-local pattern repetition was then extended to across multiple scales and demonstrated to be a strong property for natural images \cite{zontak2011internal,glasner2009super}. Several self-similarity based approaches \cite{glasner2009super,freedman2011image,singh2014super} were first proposed for image super resolution, where they restore image details by borrowing high-frequency details from self recurrences at larger scales. The idea was then explored in other restoration tasks. For example, in image denoising, its power is further strengthened by observing that noise reduces drastically at coarser scales \cite{zontak2013separating}. This motivates many advanced approaches \cite{zontak2013separating,michaeli2014blind} to restore clean signals by finding ``noisy-free" recurrences in a built image-space pyramid, yielding high-quality reconstructions. The idea of utilizing multi-scale non-local prior has achieved great successes in various restoration tasks \cite{bahat2016blind,zontak2013separating,michaeli2014blind,lotan2016needle}.

%\textbf{NL attention are bad for restoration: 1.only in-scale. }
% In a past few years, convolution neural networks for image restoration have made unprecedented progresses over conventional approaches [cite].
Recently deep neural networks trained for image restoration have made unprecedented progress. Following the importance of self-similarity prior, most recent approaches based on neural networks \cite{zhang2019residual,liu2018non} adapt non-local operations into their networks, following the \textit{non-local neural networks} \cite{wang2018non}. In a non-local block, a response is calculated as a weighted sum over all pixel-wise features on the feature map, thus it can obtain long-range information. Such a module was initially designed for high-level recognition tasks and proven to be also effective in low-level vision problems~\cite{zhang2019residual,liu2018non}. 

However, these approaches which adapt the naive self-attention module to low-level tasks have certain limitations. First, to our best knowledge, multi-scale non-local prior is never explored. It has been demonstrated in the literature that cross-scale self-similarity can bring impressive benefits for image restoration \cite{zontak2013separating,bahat2016blind,michaeli2014blind,glasner2009super}. Unlike high-level semantic features for recognition which makes not too much difference across scales, low-level features represent richer details, patterns, and textures at different scales. %Therefore, features at finer scale may contain richer information than coarser ones. 
Nevertheless, the leading non-local self-attention fails to capture the useful correspondences that occur at different scales. Second, %computing feature correlations in a pixel-wise manner brings noisy matches, 
pixel-wise matching used in the self-attention module is usually noisy for low-level vision tasks, thus reducing performance. Intuitively, enlarging the searching space raises possibility for finding better matches, but it is not true for the existing self-attention modules~\cite{liu2018non}. %The best practice only comes by restricting the search space to a very confined neighborhood \cite{liu2018non}. 
Unlike high-level feature maps where numerous dimension reduction operations are employed, image restoration networks often maintain the input spatial size. Therefore, feature is only highly relevant to a localized region, making them easily affected by noisy signals. This is in line with conventional non-local filtering, where pixel-wise matching performs much worse than block matching \cite{buades2011non}. 

%\textbf{Propose PA. benefits: 1. multi-scale 2.patch based ->more robust}
\begin{figure}[t]
    \centering
    \includegraphics[width=0.9\textwidth]{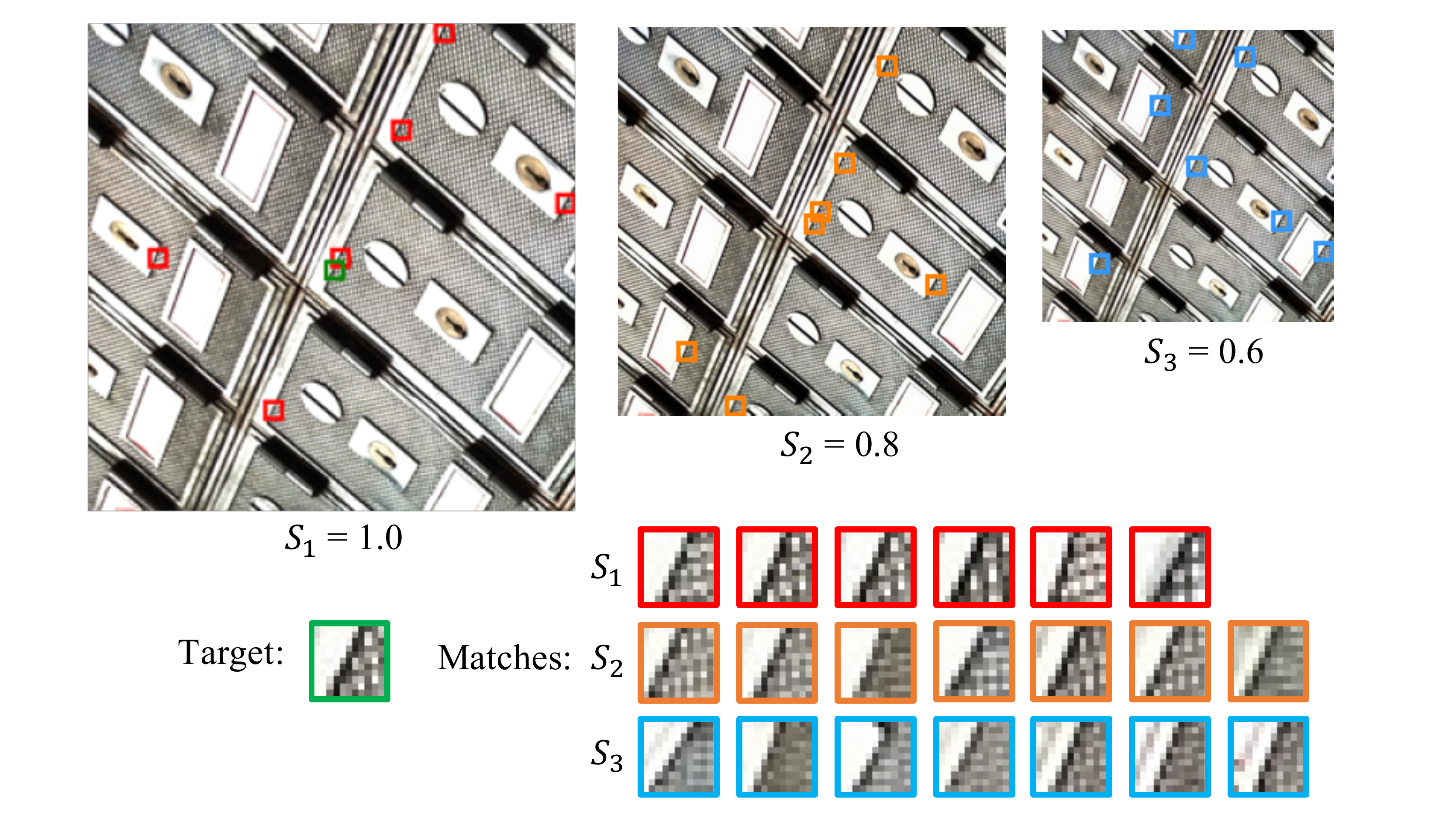}
    \caption{Visualization of most correlated patches captured by our pyramid attention. Pyramid attention exploits multi-scale self-exemplars to improve reconstruction}
    \label{fig:intro}
    \vspace{-1em}
\end{figure}
In this paper, we present a novel non-local pyramid attention as a simple and generic building block for exhaustively capturing long-range dependencies, as shown in Fig. \ref{fig:intro}. The proposed attention takes full advantages of traditional non-local operations but is designed to better accord with the nature of image restoration. Specifically, the original search space is largely extended from a single feature map to a multi-scale feature pyramid. The proposed operation exhaustively evaluates correlation among features across multiple specified scales by searching over the entire pyramid. This brings several advantages: (1) It generalizes existing non-local operation, where the original searching space is inherently covered in the lowest pyramid level. (2) The long-range dependency between relevant features of different sizes is explicitly modeled. Since the operation is fully differentiable, it can be jointly optimized with networks through back propagation. (3) Similar to traditional approaches \cite{zontak2013separating,bahat2016blind,michaeli2014blind}, one may expect noisy signals in features can be drastically reduced via rescaling to coarser pyramid level via operations like bi-cubic interpolation. This allows the network to find ``clean signal" from multi-scale correspondences. Next, we enhance the robustness of correlation measurement by involving neighboring features into computation, inspired by traditional block matching strategy. Region-to-region matching imposes additional similarity constraints on the neighborhood. As such, the module can effectively single out highly relevant correspondences while suppressing noisy ones. 

We demonstrate the power of non-local pyramid attention on various image restoration tasks: image denoising, image demosaicing, compression artifacts reduction and image super resolution. In all tasks, a single pyramid attention, which is our basic unit, can model long-range dependency without scale restriction, in a feed forward manner. With one attention block inserted into a very simple backbone network, the model achieves significantly better results than the latest state-of-the-art approach with well-engineered architecture and multiple non-local attention units. In addition, we also conduct extensive ablation studies to analyze our design choices. All these evidences demonstrate our module is a better alternative of current non-local operation and can be used as a fundamental unit in neural networks for generic image restoration.

%To fully showcase its generality against low-level vision problems, we further evaluates its effectiveness on image super resolution task. By inserting the attention module on strong baselines RCAN and EDSR, we observed considerable improvements overall all benchmark datasets. 

%#### related works ####
\section{Related Works}
\textbf{Self-similarity Prior for Image Restoration.} Self-similarity property that small patterns tend to recur within a image powers natural images with strong self-predictive ability \cite{bahat2016blind,glasner2009super, zontak2011internal}, which forms a basis for many classical image restoration methods \cite{zontak2011internal,zontak2013separating,bahat2016blind,michaeli2014blind,huang2015single}.  The initial work, non-local mean filtering \cite{buades2005non}, globally averages similar patches for image denoising. Later on,  Dabov et al \cite{dabov2007image} introduced BM3D, where repetitive patterns are grouped into 3D arrays to be jointly processed by collaborative filters. In LSSC \cite{mairal2009non}, self-similarity property is combined with sparse dictionary learning for both denoising and demosaicing. This ``fractal like" characteristic was further strengthened to \textbf{across different scales} and shown to be a very strong property for natural images \cite{glasner2009super,zontak2011internal}. To enjoy cross-scale redundancy, self-similarity based approaches were proposed for image super resolution \cite{glasner2009super,freedman2011image,huang2015single}, where high frequency information is retrieved uniquely from internal multi-scale recurrences. Observing that corruptions drop drastically at coarser scales, Zontak \cite{zontak2013separating} demonstrated that a clean version of noisy patches (99\%) exists at coarser level of the original image. This idea was developed into their denoising algorithm, which achieved promising results. The cross-scale self similarity is also of central importance for many image deblurring \cite{michaeli2014blind,bahat2017non} and image dehazing approaches \cite{bahat2016blind}.

\textbf{Non-local Attention.} Non-local attention in deep CNNs was initially proposed by Wang et al \cite{wang2018non} for video classification. In their networks, non-local units are placed on high-level, sub-sampled feature maps to compute long-range semantic correlations. By assigning weights to features at all locations, it allows the network to focus on more informative areas. Adapting non-local operation also showed considerable improvements in other high-level tasks, such as object detection \cite{cao2019gcnet}, semantic segmentation \cite{fu2019dual} and person Re-id \cite{xia2019second}. For image restoration, recent approaches, such as NLRN \cite{liu2018non}, RNAN \cite{zhang2019residual} and SAN \cite{dai2019second}, incorporate non-local operations in their networks. However, without careful modification, their performances are limited by simple single-scale correlations and further reduced by involving many ill matches during the pixel-wise feature matching in attention units. Recently, Mei et al \cite{Mei2020image} proposed a cross-scale attention and achieved state-of-the-art results in image SR. However, it can only capture correlations at two integer scales and has an additional upsampling process, making it unsuitable for general image restoration. Our work can be considered as a significant extension and methodological generalization of \cite{Mei2020image}.

\textbf{Deep CNNs for Image Restoration.} Adopting deep-CNNs for image restoration has shown evident improvements by embracing their representative power. In the early work, Vincent et al \cite{vincent2008extracting} proposed to use stacked auto-encoder for image denoising. Later, ARCNN was introduced by Dong et al \cite{dong2015compression} for compression artifacts reduction. Zhang et al \cite{zhang2017beyond} proposed DnCNN for image denosing, which uses advanced techniques like residual learning and batch normalization to boost performance. In IRCNN \cite{zhang2017learning}, a learned set of CNNs are used as denoising prior for other image restoration tasks. For image super resolution, extensive efforts have been spent into designing advanced architectures and learning methods, such as progressive super resolution \cite{lai2017deep}, residual \cite{lim2017enhanced} and dense connection \cite{zhang2018residual}, back-projection \cite{haris2018deep}, scale-invariant convolution \cite{fan2019scale} and channel attention \cite{zhang2018image}. Recently, most state-of-the-art approaches \cite{liu2018non,zhang2019residual,dai2019second} incorporate non-local attention into networks to further boost representation ability. Although extensive efforts have been made in architectural engineering, existing methods relying on convolution and non-local operation can only exploit information at a same scale.

%#### attention ####
\section{Pyramid Attention Networks}
Both convolution operation and non-local attention are restricted to same-scale information. In this section, we introduce the novel pyramid attention, which can deal with non-local dependency across multiple scales, as a generalization of non-local operations.
%The proposed pyramid attention is a generalization of the non-local operations, which are designed to deal with non-local dependency across multiple scales. In this section, we will discuss this operation in depth.     
\subsection{Formal Definition}
\begin{figure*}[t] 
  \begin{center}
    \begin{tabular}[c]{ccc}
      \includegraphics[ width=.33\textwidth]{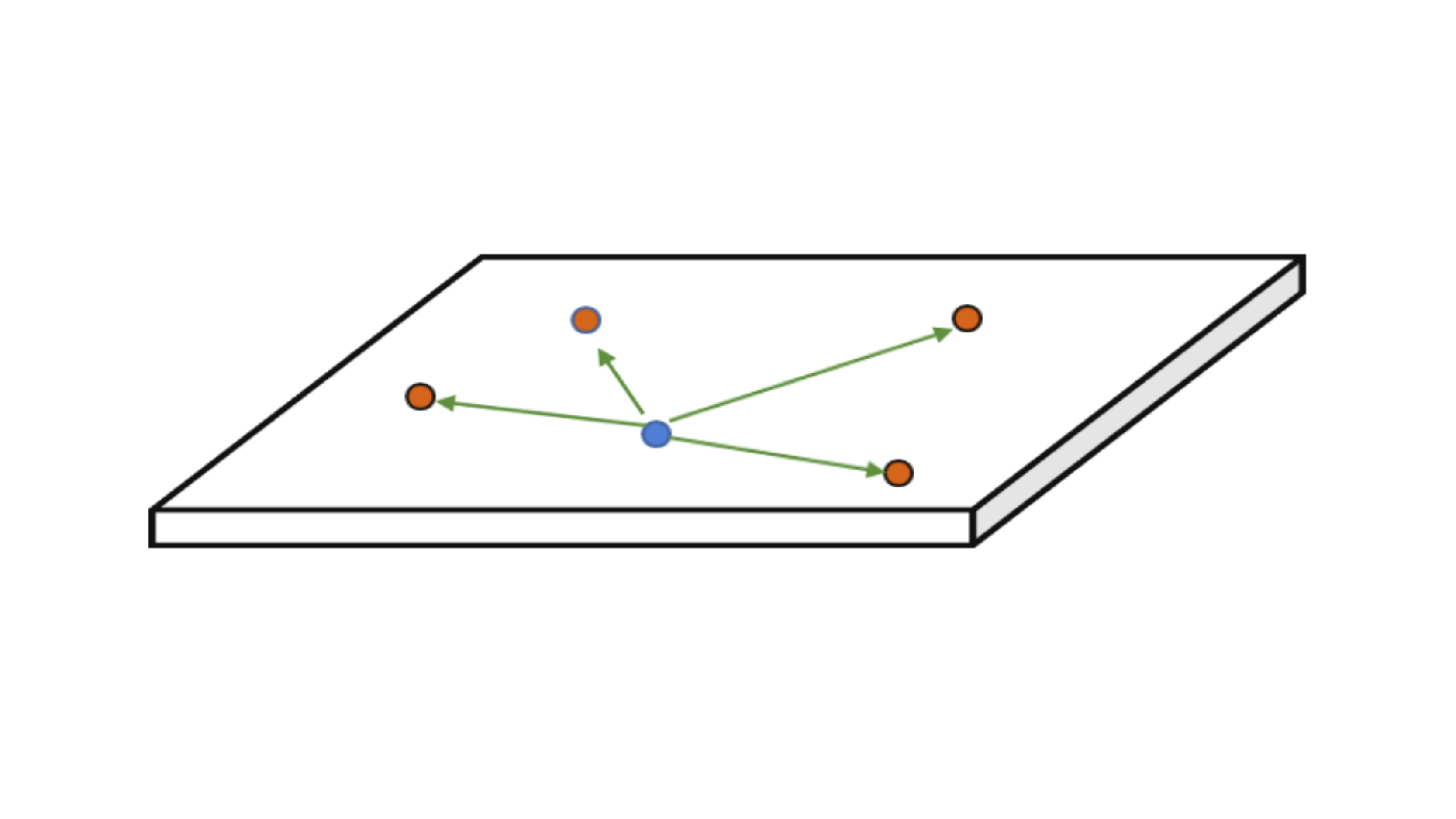} 
      &
        \includegraphics[width=.33\textwidth]{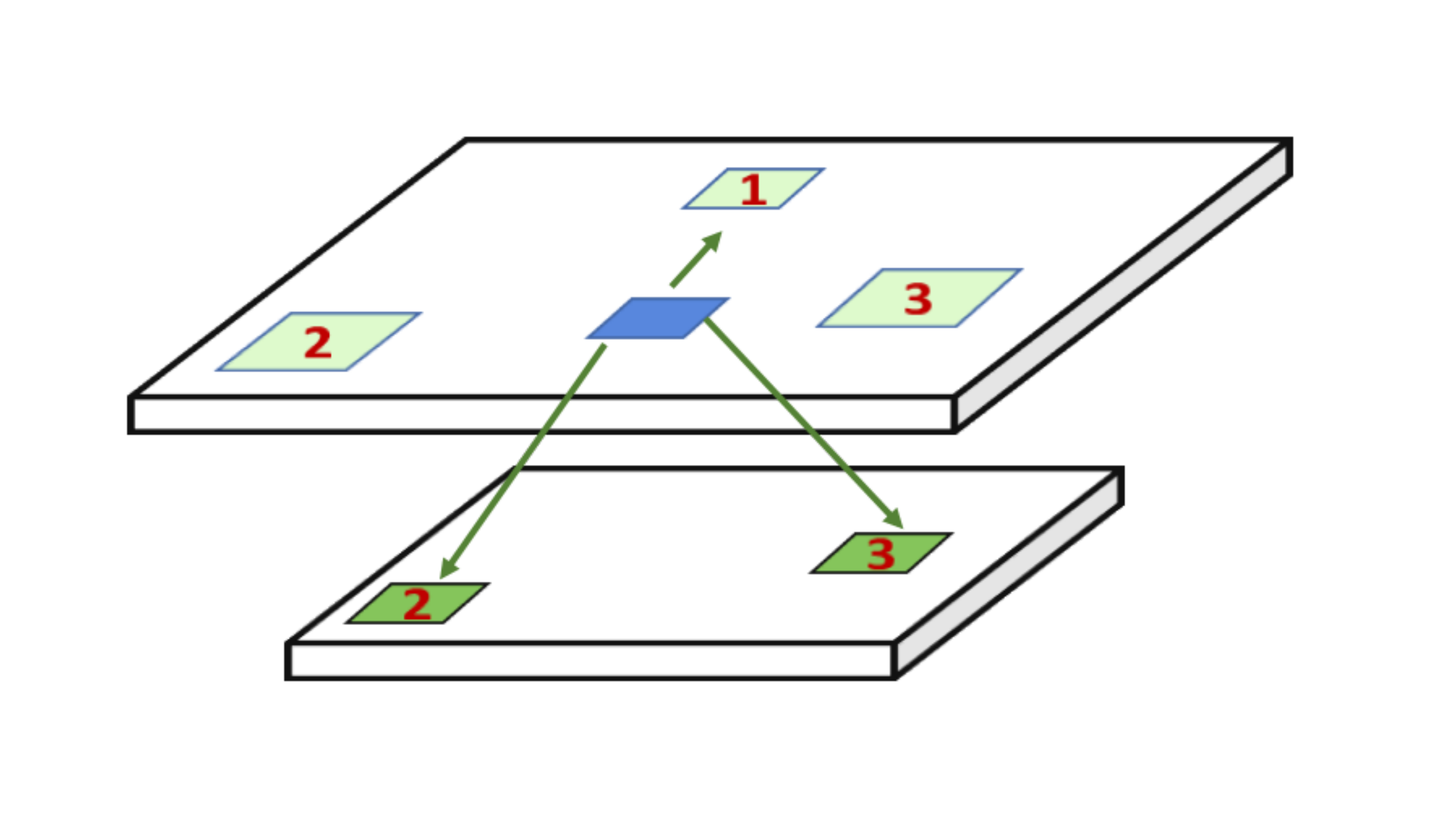} 
      &
        \includegraphics[ width=.33\textwidth]{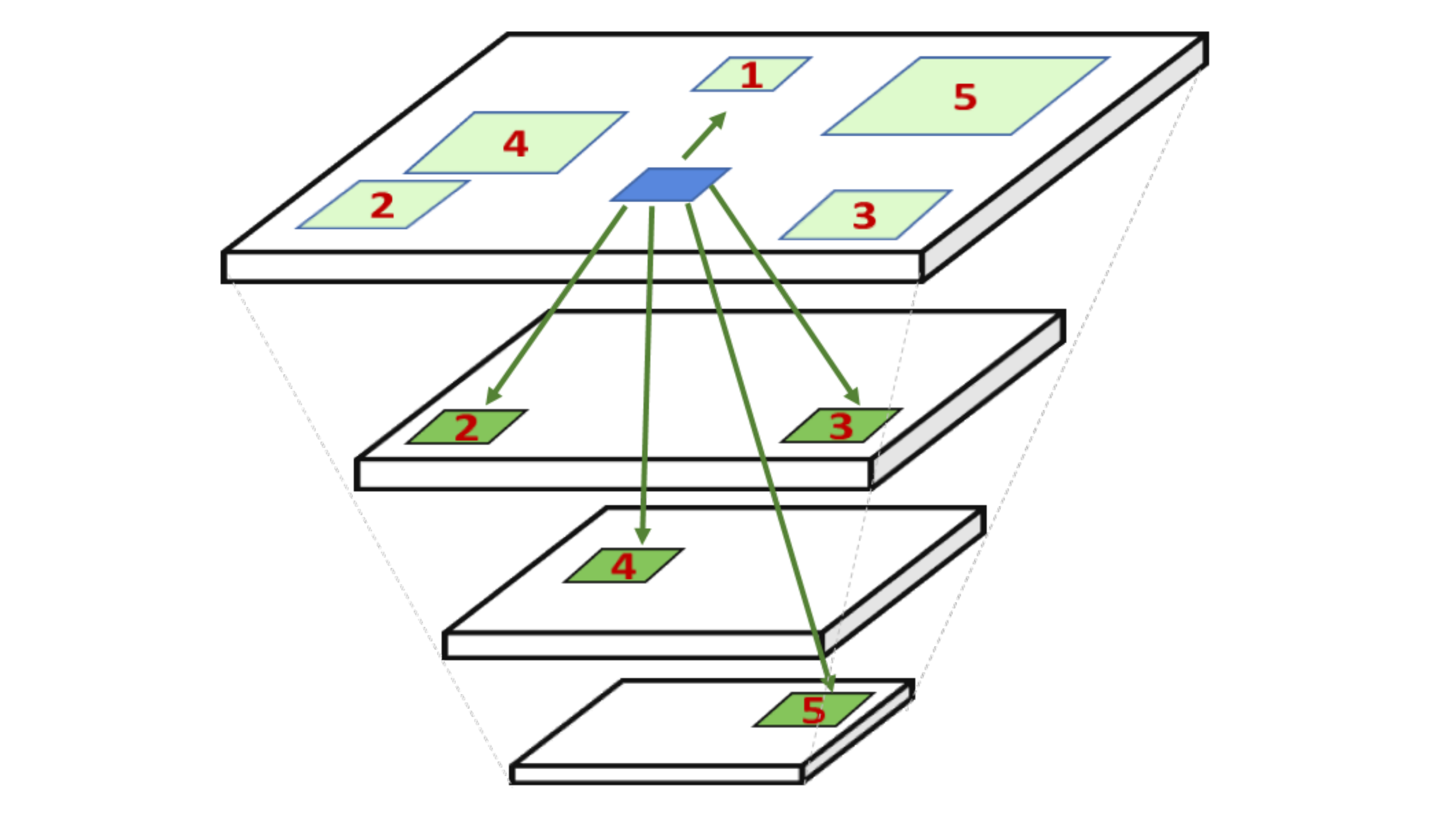} 
        \\
      {\small (a)  Non-local attention}
      &{\small (b) Scale agnostic attention}
      &{\small (c) Pyramid attention}\\
    \end{tabular}\vspace{0.2em}
    \caption{Comparison of attentions. (a) classic self-attention computes pair-wise feature correlation at same scale. (b) Scale agnostic attention augments (a) to capture correspondences at one additional scale. (c) Pyramid attention generalizes (a) and (b) by modeling multi-scale non-local dependency   
    }
    \label{fig:attention}\vspace{-1em}
  \end{center}
\end{figure*}

Non-local attention calculates a response by averaging features over an entire image, as shown in Fig. \ref{fig:attention} (a). Formally, given an input feature map $x$, this operation is defined as:
\begin{equation}
    y^{i} = \frac{1}{\mathcal{\sigma}(x)}\sum_{j}\phi(x^{i},x^{j})\theta(x^{j}),
    \label{eq:1}
\end{equation}
where $i$, $j$ are index on the input $x$ and output $y$ respectively. The function $\phi$ computes pair-wise affinity between two input features. $\theta$ is a feature transformation function that generates a new representation of $x^{j}$. The output response $y^{i}$ obtains information from all features by explicitly summing over all positions and is normalized by a scalar function $\mathcal{\sigma}(x)$. While the above operation manages to capture long-range correlation, information is extracted at a single scale. %Since features on a input map cover exactly same range of information, 
As a result, it fails to exploit relationships to many more informative areas of distinctive spatial sizes. %, where spatial sizes differ from the current one. 

To break this scale constraint, we propose pyramid attention (Fig. \ref{fig:attention} (c)), which captures correlations across scales. In pyramid attention, affinities are computed between a target feature and regions. Therefore, a response feature is a weighted sum over multi-scale correspondences within the input map. Formally, given a series of scale factor $S = \{1, s_{1}, s_{2},..., s_{n}\}$, pyramid attention can be expressed as
\begin{equation}
        y^{i} = \frac{1}{\sigma(x)}\sum_{s\in{S}}\sum_{j}\phi(x^{i},x^{j}_{\delta(s)}) \theta(x^{j}_{\delta(s)}). 
\end{equation}
Here $\delta(s)$ represents a $s^{2}$ neighborhood centred at index $j$ on input $x$. 

In other words, pyramid attention behaves in a non-local multi-scale way by explicitly processing larger regions with sizes specified by scale pyramid $S$ at all position $j$. Note that when only a single scale factor $s = 1$ is specified, the proposed attention degrades to current non-local operation. Hence, our approach is a more generic operation that allows the network to fully enjoy the predictive power of natural images.

Finding a generic solution, which models cross-scale relationships, is a non-trivial problem and requires carefully engineering. In the following section, we first address the non-local operation between two scales and then extend it to pyramid scales. 
\subsection{Scale Agnostic Attention} Given an extra scale factor $s$, how to evaluate the correlation between $x^{j}$ and $x^{j}_{\delta(s)}$ and aggregate information from $x^{j}_{\delta(s)}$ to form $y^{i}$ are two key steps. Here, the major difficulty comes from misalignment in their spatial dimensions. Common similarity measurements, such as dot product and embedded Gaussian, only accept features with identical dimensions, thus are infeasible in this case. 

To mitigate the above problem, we propose to squeeze the spatial information of $x^{j}_{\delta(s)}$ into a single region descriptor. This step is conducted by down-scaling the region $x^{j}_{\delta(s)}$ in a pixel feature $z^{j}$. As we need search over the entire feature map, we can therefore directly down-scale the original input $x$ (H$\times$W) to obtain a descriptor map $z$ ($\frac{H}{s}\times\frac{W}{s}$). The correlation between $x^{i}$ and $x^{j}_{\delta(s)}$ is then represented by $x^{i}$ and the region descriptor $z^{j}$. Formally, scale agnostic attention (Fig. \ref{fig:attention} (b)) is formulated as
\begin{align} \label{eq:4}
    y^{i} = \frac{1}{\sigma(x,z)}\sum_{j}\phi(x^{i},z^{j})\theta(z^{j}),
\end{align}
where $z = x\downarrow s$. 

This operation brings additional advantages. As discussed in Section \ref{sec:1}, downscaling regions into coarser descriptors reduces noisy levels. On the other hand, since the cross-scale recurrence represents a similar content, the structure information will be still well-preserved after down-scaling. Combing these two facts, region descriptors can serve as a ``cleaner version" of the target feature and a better alternative of noisy patch matches at the original scale.  
%not pyramid, only one scale to another
% general form matches with scale factor s
% how to achieve, down-sample, to obtain same size
% effect noisy averaging, may be better alternative for in-scale candidates 

\subsection{Pyramid Attention}
% how pyramid formed?
% level 
To make full use of self-predictive power, the scale agnostic attention can be extended to pyramid attention, which computes correlations across multiple scales. In such units, pixel-region correspondences are captured over an entire feature pyramid. Specifically, given a series of scales $S=\{1, s_{1}, s_{2},.., s_{n}\}$, it forms a feature pyramid $\mathcal{F} = \{F_{1}, F_{2},..., F_{n}\}$, where $F_{i}$ ($\frac{H}{s_{i}}\times\frac{W}{s_{i}}$) is a region descriptor map of the input $x$, obtained by down-scaling operation. In such case, the correlations between any pyramid levels and the original input $x$ can be seen as a scale agnostic attention. Therefore, the pyramid attention is defined as:
\begin{equation}
    y^{i} = \frac{1}{\sigma(x,\mathcal{F})} \sum_{z\in \mathcal{F}} \sum_{j\in z}\phi(x^{i},z^{j})\theta(z^{j}).
\end{equation}
The cross-scale modeling ability is due to the fact that region descriptor $z_{i}$ at different levels summarizes information over regions of various sizes. When they are copied back to original position $i$, non-local multi-scale information is fused together to form a new response, which intuitively contains richer and more faithful information than the matches from a single scale.

\subsection{Instantiation} 
\textbf{Choices of $\phi$, $\theta$ and $\sigma$.} There are many well-explored choices for pair-wise function $\phi$ \cite{wang2018non,liu2018non}, such as Gaussian, embedded Gaussian, dot pot and feature concatenation. In this paper, we use embedded Gaussian to follow previous best practices \cite{liu2018non}: $\phi(x^{i},z^{j}) = e^{f(x^{i})^{T}g(z^{j})}$, where $f(x^{i})=W_{f}x^{i}$ and $g(z^{j})=W_{g}z^{j}$.

For feature transformation function $\theta$, we use a simple linear embedding: $\theta = W_{\theta}z^{j}$. Finally, we set $\sigma(x,\mathcal{F})= \sum_{z\in \mathcal{F}}\sum_{j\in z}\phi(x^{i},z^{j})$. By specifying above instantiations, the term $\frac{1}{\sigma(x,F)}\sum_{x\in\mathcal{F}}\Phi(x^{i},z^{j})$ is equivalent to softmax over all possible positions in the pyramid.  

\textbf{Patch based region-to-region attention} As discussed in Section \ref{sec:1}, information contained in features (for image restoration tasks) is very localized. Consequently, the matching process is usually affected by noisy signals. Previous approach relieves this problem by restriction search space to local region \cite{liu2018non}. However, this also prevents them from finding better correspondences that are far away from current position. 

To improve the robustness during matching, we impose extra neighborhood similarity, which is in line with classical non-local filtering \cite{buades2005non}. As such, the pyramid attention (eq. \ref{eq:4}) is expressed as:
\begin{equation}
     y^{i} = \frac{1}{\sigma(x,\mathcal{F})} \sum_{z\in \mathcal{F}} \sum_{j\in z}\phi(x^{i}_{\delta{(r)}},z^{j}_{\delta{(r)}})\theta(z^{j}),
\end{equation}
where the neighborhood is specified by $\delta(r)$. This add a stronger constraint on matching content that two features are highly correlated if and only if their neighborhood are highly similar as well. The block-wise matching allows the network to pay more attention on relevant areas while suppressing unrelated ones.

\begin{figure}[t] \label{ResNet}
\centering
\includegraphics[ width=1\textwidth]{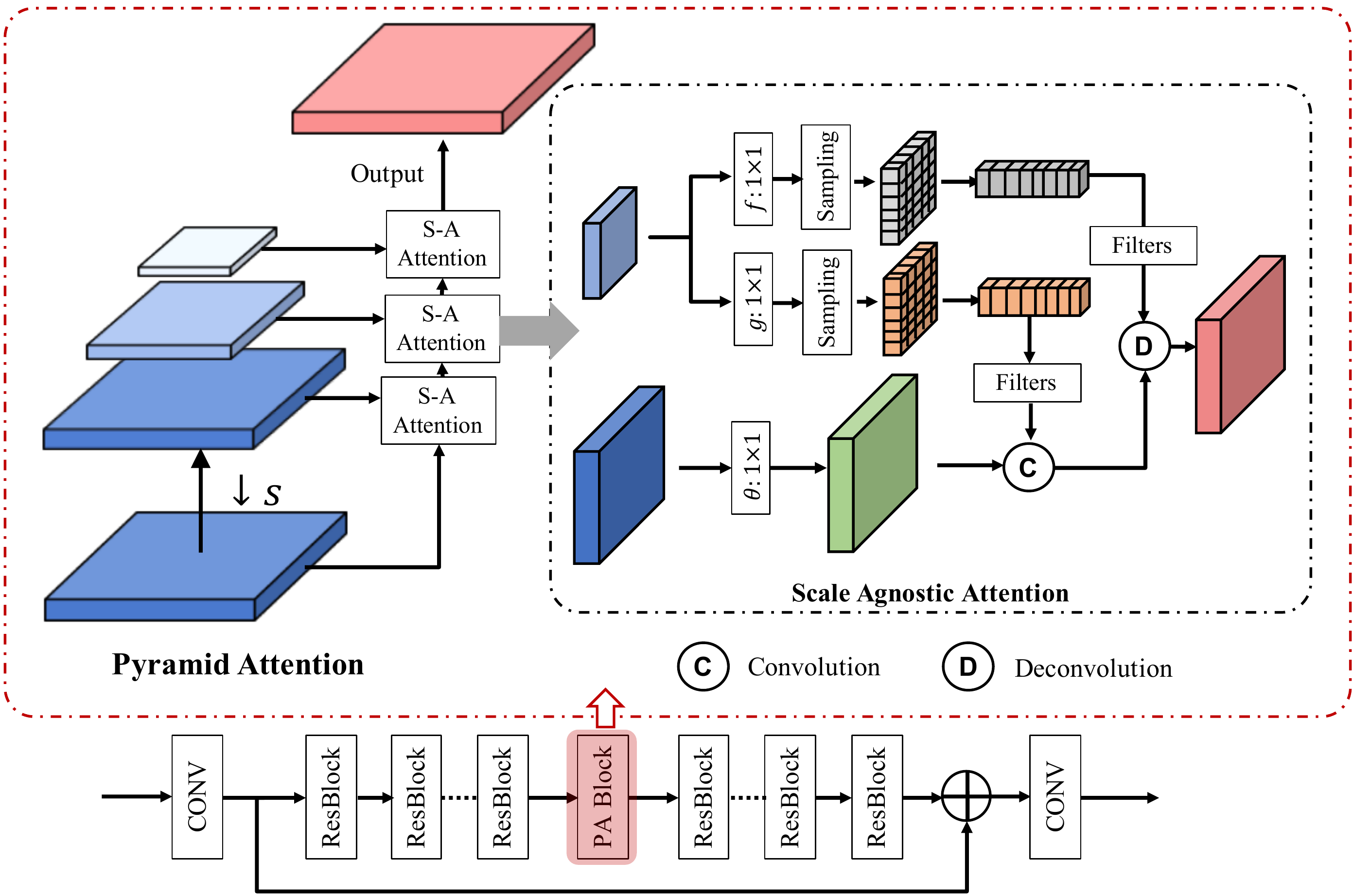}
\vspace{-1em}
\caption{PANet with the proposed pyramid attention (PA). Pyramid attention captures multi-scale correlation by consecutively computing Scale Agnostic (S-A) attention}
\label{fig:PANet}
\vspace{-4mm}
\end{figure}

\textbf{Implementation.} The proposed pyramid attention is implemented using basic convolution and deconvolution operations, as shown in Fig. \ref{fig:PANet}. In practice, softmax matching scores can be expressed as convolution over the input $x$ using $r\times r$ patches extracted from the feature pyramid. To obtain a final response, we extract patches from the transformed feature map (by $\theta$) to conduct a deconvolution over the matching score. Note that the proposed operation is fully convolutional, differential and accept any input resolutions, which can be flexibly embedded into many standard architectures. 

%\subsection{Exemplar: PANet}
\subsection{PANet: Pyramid Attention Networks}

To show the effectiveness of our pyramid attention, we choose a simple ResNet as our backbone without any architectural engineering. The proposed image restoration network is illustrated in Fig. \ref{fig:PANet}. We remove batch normalization in each residual block, following the practice in \cite{lim2017enhanced}. Similar to many restoration networks, we add a global pathway from the first feature to the last one, which encourages the network bypass low frequency information. We inset a single pyramid attention in the middle of the network.

Given a set of N-paired training images $I^{k}_{LQ}$-$I^{k}_{HQ}$, we optimize the $L1$ reconstruction loss between $I^{k}_{HQ}$ and $I^{K}_{RQ} =R(I^{K}_{LQ},\Theta)$,
\begin{equation}
    \mathcal{L}_{1} = \frac{1}{N}\sum_{i=1}^{N}\|I^{K}_{RQ}-I^{K}_{HQ}\|,
\end{equation}
where $R$ represents the entire PANet and $\Theta$ is a set of learnable parameters.

%#### experiments ####
\section{Experiments}\label{sec:4}
\subsection{Datasets and Evaluation Metrics}
The proposed pyramid attention and PANet are evaluated on major image restoration tasks: image denoising, demosaicing and compression artifacts reduction. For fair comparison, we follow the setting specified by RNAN~\cite{zhang2019residual} for image denoising, demosaicing, and compression artifacts reduction. We use DIV2K~\cite{timofte2017ntire} as our training set, which contains 800 high quality images. We report results on standard benchmarks using PSNR and/or SSIM~\cite{wang2004image}.
\subsection{Implementation Details}
For pyramid attention, we set the scale factors $S = \{1.0, 0.9, 0.8, 0.7, 0.6\}$, so that we construct a 5 level feature pyramid within the attention block. To build the pyramid, we use simple Bicubic interpolation to rescale feature maps. While computing correlations, we use $3\times 3$ small patches centered at target features. The proposed PANet contains 80 residual blocks with one pyramid attention module inserted after the 40-$th$ block. All features have 64 channels, except for those used in embedded Gaussian, where the channel number is reduced to 32.

During training, each mini-batch consists of 16 patches with size $48\times 48$. We augment training images using vertical/horizontal flipping and random rotation of ${90}^{\circ}$, ${180}^{\circ}$, and ${270}^{\circ}$. The model is optimized by Adam optimizer with $\beta_1=0.9$, $\beta_2=0.999$, and $\epsilon=10^{-8}$. The learning rate is initialized to $10^{-4}$ and reduced to a half after every 200 epochs. Our model is implemented using PyTorch~\cite{paszke2017automatic} and trained on Nvidia TITANX GPUs. 

\begin{table}[t]
%\scriptsize
%\footnotesize
%\small
%\normalsize
\center
\begin{center}
\caption{Quantitative evaluation of state-of-the-art approaches on color image denoising. Best results are \textbf{highlighted}}
\label{tab:results_psnr_denoise_rgb}
\vspace{-2mm}
%\begin{tabular*}{75.8mm}{@{\extracolsep{-0.99mm}}cccccccccc|c|c|c|c|c|c|c|c|c|c|c|c|c|c|c|c|c|c|}
\resizebox{1.0\columnwidth}{!}{
\begin{tabular}{|l|c|c|c|c|c|c|c|c|c|c|c|c|c|c|c|c|}
\hline
\multirow{2}{*}{Method} &  \multicolumn{4}{c|}{Kodak24} &  \multicolumn{4}{c|}{BSD68} &  \multicolumn{4}{c|}{Urban100}   
\\
%\hline
\cline{2-13}
 & 10 & 30 & 50 & 70 & 10 & 30 & 50 & 70 & 10 & 30 & 50 & 70 
\\
\hline
\hline
%\multirow{4}{*}{LIVE1}
CBM3D
& 36.57
 & 30.89
  & 28.63
   & 27.27
    & 35.91
     & 29.73
      & 27.38
       & 26.00
        & 36.00
         & 30.36
          & 27.94
           & 26.31
           
\\
\hline
TNRD
& 34.33
 & 28.83
  & 27.17
   & 24.94
    & 33.36
     & 27.64
      & 25.96
       & 23.83
        & 33.60
         & 27.40
          & 25.52
           & 22.63
           
\\
\hline
RED
& 34.91
 & 29.71
  & 27.62
   & 26.36
    & 33.89
     & 28.46
      & 26.35
       & 25.09
        & 34.59
         & 29.02
          & 26.40
           & 24.74
           
\\
\hline
DnCNN
& 36.98
 & 31.39
  & 29.16
   & 27.64
    & 36.31
     & 30.40
      & 28.01
       & 26.56
        & 36.21
         & 30.28
          & 28.16
           & 26.17
            
\\
\hline
MemNet
& N/A
 & 29.67
  & 27.65
   & 26.40
    & N/A
     & 28.39
      & 26.33
       & 25.08
        & N/A
         & 28.93
          & 26.53
           & 24.93
           
\\
\hline
IRCNN
& 36.70
 & 31.24
  & 28.93
   & N/A
    & 36.06
     & 30.22
      & 27.86
       & N/A
        & 35.81
         & 30.28
          & 27.69
           & N/A
           
\\
\hline
FFDNet
& \underline{36.81}
 & 31.39
  & 29.10
   & 27.68
    & 36.14
     & 30.31
      & 27.96
       & 26.53
        & 35.77
         & 30.53
          & 28.05
           & 26.39
           
\\
\hline
RNAN
& \underline{37.24}
 & \underline{31.86}
  & \underline{29.58}
   & \underline{28.16}
    & \underline{36.43}
     & \underline{30.63}
      & \underline{28.27}
       & \underline{26.83}
        & \underline{36.59}
         & \underline{31.50}
          & \underline{29.08}
           & \underline{27.45}
           
\\
\hline
PANet
& \textbf{37.35}
 & \textbf{31.96}
  & \textbf{29.65}
   & \textbf{28.20}
    & \textbf{36.50}
     & \textbf{30.70}
      & \textbf{28.33}
       & \textbf{26.89}
        & \textbf{36.80}
         & \textbf{31.87}
          & \textbf{29.47}
           & \textbf{27.87}
           \\
\hline         
\end{tabular}
}
\end{center}
%\vspace{-5mm}
\end{table}
\begin{figure}[ht]
%\newlength-4mm
%\setlength{-4mm}{-0.4cm}
\scriptsize
\centering
\begin{tabular}{cc}
% % one row
\hspace{-0.2cm}
\begin{adjustbox}{valign=t}
\begin{tabular}{c}
\includegraphics[width=0.2035\textwidth,height=0.245\textwidth]{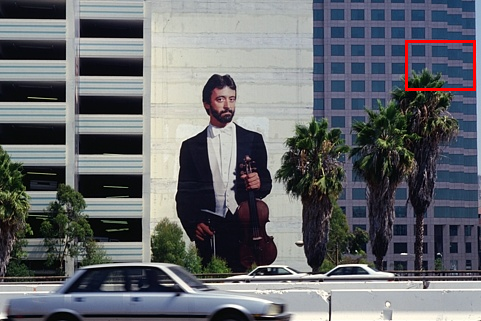}
\\
BSD68: 119082 
\end{tabular}
\end{adjustbox}
\hspace{-0.23cm}
\begin{adjustbox}{valign=t}
\begin{tabular}{cccccc}
\includegraphics[width=0.15\textwidth]{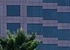} \hspace{-1.5mm} &
\includegraphics[width=0.15\textwidth]{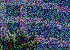} \hspace{-1.5mm} &
\includegraphics[width=0.15\textwidth]{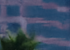} \hspace{-1.5mm} &
\includegraphics[width=0.15\textwidth]{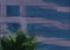} \hspace{-1.5mm} &
\includegraphics[width=0.15\textwidth]{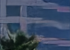} \hspace{-1.5mm} 
\\
HQ \hspace{-1.5mm} &
Noisy ($\sigma$=50) \hspace{-1.5mm} &
TNRD~\cite{chen2017trainable} \hspace{-1.5mm} &
RED~\cite{mao2016image} \hspace{-1.5mm} &
DnCNN~\cite{zhang2017beyond} \hspace{-1.5mm} 
\\
\includegraphics[width=0.15\textwidth]{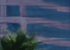} \hspace{-1.5mm} &
\includegraphics[width=0.15\textwidth]{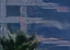} \hspace{-1.5mm} &
\includegraphics[width=0.15\textwidth]{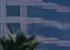} \hspace{-1.5mm} &
\includegraphics[width=0.15\textwidth]{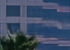} \hspace{-1.5mm} &
\includegraphics[width=0.15\textwidth]{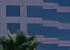} \hspace{-1.5mm}  
\\ 
MemNet~\cite{tai2017memnet} \hspace{-1.5mm} &
IRCNN~\cite{zhang2017learning} \hspace{-1.5mm} &
FFDNet~\cite{zhang2017ffdnet} \hspace{-1.5mm} &
RNAN~\cite{zhang2019residual}  \hspace{-1.5mm} &
PANet \hspace{-1.5mm} 
\\
\end{tabular}
\end{adjustbox}
\vspace{2mm}
\\
% % one row
\hspace{-0.2cm}
\begin{adjustbox}{valign=t}
\begin{tabular}{c}
\includegraphics[width=0.2035\textwidth,height=0.245\textwidth]{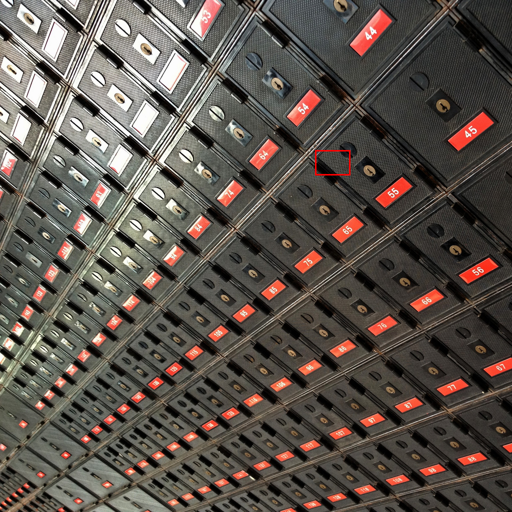}
\\
Urban100: img006
\end{tabular}
\end{adjustbox}
\hspace{-0.23cm}
\begin{adjustbox}{valign=t}
\begin{tabular}{cccccc}
\includegraphics[width=0.15\textwidth]{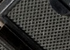} \hspace{-1.5mm} &
\includegraphics[width=0.15\textwidth]{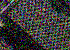} \hspace{-1.5mm} &
\includegraphics[width=0.15\textwidth]{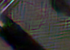} \hspace{-1.5mm} &
\includegraphics[width=0.15\textwidth]{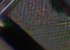} \hspace{-1.5mm} &
\includegraphics[width=0.15\textwidth]{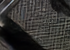} \hspace{-1.5mm} 
\\
HQ \hspace{-1.5mm} &
Noisy ($\sigma$=50) \hspace{-1.5mm} &
TNRD~\cite{chen2017trainable} \hspace{-1.5mm} &
RED~\cite{mao2016image} \hspace{-1.5mm} &
DnCNN~\cite{zhang2017beyond} \hspace{-1.5mm} 
\\
\includegraphics[width=0.15\textwidth]{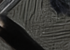} \hspace{-1.5mm} &
\includegraphics[width=0.15\textwidth]{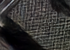} \hspace{-1.5mm} &
\includegraphics[width=0.15\textwidth]{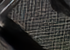} \hspace{-1.5mm} &
\includegraphics[width=0.15\textwidth]{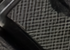} \hspace{-1.5mm} &
\includegraphics[width=0.15\textwidth]{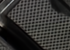} \hspace{-1.5mm}  
\\ 
MemNet~\cite{tai2017memnet} \hspace{-1.5mm} &
IRCNN~\cite{zhang2017learning} \hspace{-1.5mm} &
FFDNet~\cite{zhang2017ffdnet} \hspace{-1.5mm} &
RNAN~\cite{zhang2019residual}  \hspace{-1.5mm} &
PANet  \hspace{-1.5mm} 

\\
\end{tabular}
\end{adjustbox}
\vspace{2mm}
\\
\hspace{-0.3cm}
\begin{adjustbox}{valign=t}
\begin{tabular}{c}
\includegraphics[width=0.2035\textwidth,height=0.27\textwidth]{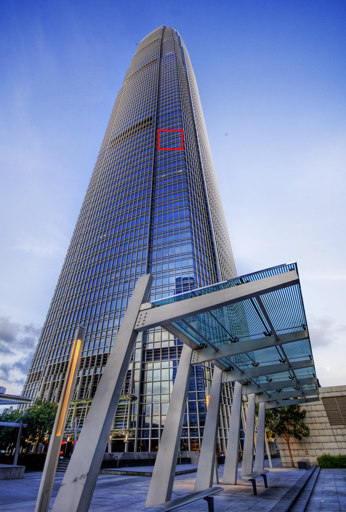}
\\
Urban100: img046
\end{tabular}
\end{adjustbox}
\hspace{-0.23cm}
\begin{adjustbox}{valign=t}
\begin{tabular}{cccccc}
\includegraphics[width=0.15\textwidth]{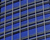} \hspace{-1.5mm} &
\includegraphics[width=0.15\textwidth]{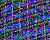} \hspace{-1.5mm} &
\includegraphics[width=0.15\textwidth]{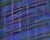} \hspace{-1.5mm} &
\includegraphics[width=0.15\textwidth]{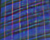} \hspace{-1.5mm} &
\includegraphics[width=0.15\textwidth]{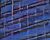} \hspace{-1.5mm} 
\\
HQ \hspace{-1.5mm} &
Noisy ($\sigma$=50) \hspace{-1.5mm} &
TNRD~\cite{chen2017trainable} \hspace{-1.5mm} &
RED~\cite{mao2016image} \hspace{-1.5mm} &
DnCNN~\cite{zhang2017beyond} \hspace{-1.5mm} 
\\
\includegraphics[width=0.15\textwidth]{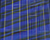} \hspace{-1.5mm} &
\includegraphics[width=0.15\textwidth]{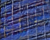} \hspace{-1.5mm} &
\includegraphics[width=0.15\textwidth]{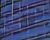} \hspace{-1.5mm} &
\includegraphics[width=0.15\textwidth]{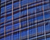} \hspace{-1.5mm} &
\includegraphics[width=0.15\textwidth]{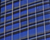} \hspace{-1.5mm}  
\\ 
MemNet~\cite{tai2017memnet} \hspace{-1.5mm} &
IRCNN~\cite{zhang2017learning} \hspace{-1.5mm} &
FFDNet~\cite{zhang2017ffdnet} \hspace{-1.5mm} &
RNAN~\cite{zhang2019residual}  \hspace{-1.5mm} &
PANet \hspace{-1.5mm} 

\\
\end{tabular}
\end{adjustbox}
\end{tabular}
\vspace{-2mm}
\caption{Visual comparison for color image denoising with noise level $\sigma$ = 50}
\label{fig:result_DN_RGB_N50_main}
\vspace{-4mm}
\end{figure}

\subsection{Image Denoising}
Following RNAN~\cite{zhang2019residual}, PANet is evaluated on standard benchmarks for image denoising: Kodak24~(\href{http://r0k.us/graphics/kodak/}{http://r0k.us/graphics/kodak/}), BSD68~\cite{martin2001database}, and Urban100~\cite{huang2015single}. We create noisy images by adding AWGN noises with $\sigma = 10, 30, 50, 70$. We compare our approach with 8 state-of-the-art methods: CBM3D~\cite{dabov2007color}, TNRD~\cite{chen2017trainable}, RED~\cite{mao2016image}, DnCNN~\cite{zhang2017beyond}, MemNet~\cite{tai2017memnet}, IRCNN~\cite{zhang2017learning}, FFDNet~\cite{zhang2017ffdnet}, and RNAN~\cite{zhang2019residual}.

As shown in Table \ref{tab:results_psnr_denoise_rgb}, PANet achieved best results on almost all datasets and noisy levels. In particular, our approach yielded better results than prior state-of-the-art RNAN, which has well-engineered network and multiple non-local attention blocks. These results show that, even with only one additional pyramid attention, a simple ResNet can significantly boost restoration quality. One may notice that PANet performs significantly well on Urban100 datasets, with more than 0.3 dB improvements over RNAN on all noisy levels. This is because pyramid attention allows the network to explicitly capture abundant cross-scale self-exemplars in urban scenes. In contrast, traditional non-local attention fails to explore those multi-scale relationships.

We further present qualitative evaluations on BSD68 and Urban100. The results are shown in Fig. \ref{fig:result_DN_RGB_N50_main}.  By relying on a single learned pyramid attention, PANet managed to produce the most faithful restoration results than others.

\begin{table}[thbp]\setlength{\tabcolsep}{6pt} \label{tab: demosaicing}
\scriptsize
%\footnotesize
%\small
%\normalsize
\center
\begin{center}
%\vspace{-5mm}
\caption{Quantitative evaluation of state-of-the-art approaches on color image demosaicing. Best results are \textbf{highlighted}}
\label{tab:results_psnr_demosaic_rgb}
\vspace{-2mm}
%\begin{tabular*}{75.8mm}{@{\extracolsep{-0.99mm}}cccccccccc|c|c|c|c|c|c|c|c|c|c|c|c|c|c|c|c|c|c|}
%\resizebox{0.9\columnwidth}{!}{
\begin{tabular}{|l|c|c|c|c|c|c|c|c|c|c|c|c|c|c|c|c|}
\hline
\multirow{2}{*}{Method} &  \multicolumn{2}{c|}{McMaster18} &  \multicolumn{2}{c|}{Kodak24} &  \multicolumn{2}{c|}{BSD68} &  \multicolumn{2}{c|}{Urban100}   
\\
%\hline
\cline{2-9}
 & PSNR & SSIM & PSNR & SSIM & PSNR & SSIM & PSNR & SSIM 
\\
\hline
\hline
Mosaiced
& 9.17 & 0.1674 & 8.56 & 0.0682 & 8.43 & 0.0850 & 7.48 & 0.1195        
\\
\hline
IRCNN
& 37.47 & 0.9615 & 40.41 & 0.9807 & 39.96 & 0.9850 & 36.64 & 0.9743         
\\
\hline
RNAN
& \underline{39.71} & \underline{0.9725} & \underline{43.09} & \underline{0.9902} & \underline{42.50} & \underline{0.9929} & \underline{39.75} & \underline{0.9848}        
\\
\hline
PANet
&\textbf{40.00} & \textbf{0.9737} &\textbf{43.29} & \textbf{0.9905} & \textbf{42.86} & \textbf{0.9933} & \textbf{40.50} & \textbf{0.9854} 
\\
\hline         
\end{tabular}
%}
\end{center}
%\vspace{-6mm}
\end{table}

\subsection{Image Demosaicing}
For image demosaicing, we conduct evaluations on Kodak24, McMaster~\cite{zhang2017learning}, BSD68, and Urban100, following settings in RNAN~\cite{zhang2019residual}. %Due to the limited space, we only 
We compare our approach with recent state-of-the-arts IRCNN~\cite{zhang2017learning} and RNAN~\cite{zhang2019residual}. As shown in Table \ref{tab:results_psnr_demosaic_rgb}, mosaic corruption significantly reduced image quality in terms of PSNR and SSIM. RNAN and IRCNN could remove these corruptions to some degree and lead to relatively high-quality restoration. Our approach yields the best reconstruction, demonstrating advantages of exploiting multi-scale correlations.

\begin{figure}[t]
%\newlength-4mm
%\setlength{-4mm}{-0.4cm}
\scriptsize
\centering
\begin{tabular}{cc}
% % one row
\hspace{-0.2cm}
\begin{adjustbox}{valign=t}
\begin{tabular}{c}
\includegraphics[width=0.2035\textwidth,height=0.15\textwidth]{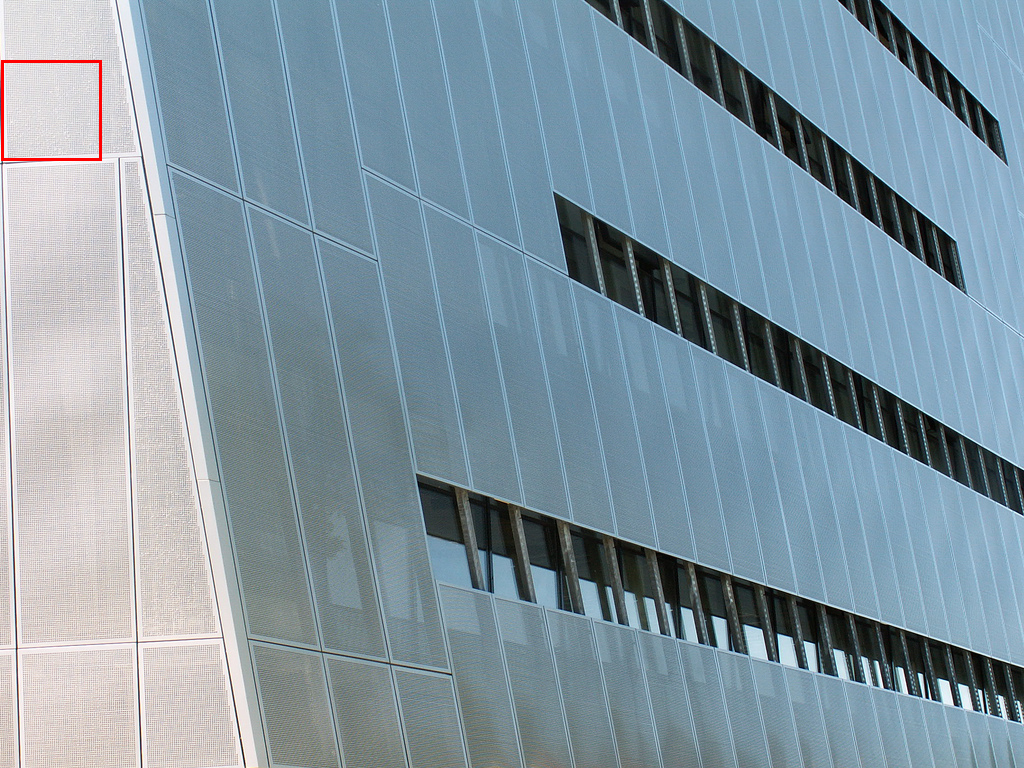}
\\
Urban100: img\_026
\end{tabular}
\end{adjustbox}
\hspace{-0.23cm}
\begin{adjustbox}{valign=t}
\begin{tabular}{cccccc}
\includegraphics[width=0.15\textwidth]{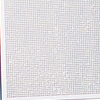} \hspace{-1mm} &
\includegraphics[width=0.15\textwidth]{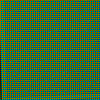} \hspace{-1mm} &
\includegraphics[width=0.15\textwidth]{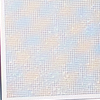} \hspace{-1mm} &
\includegraphics[width=0.15\textwidth]{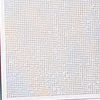} \hspace{-1mm} &
\includegraphics[width=0.15\textwidth]{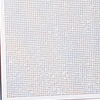} \hspace{-1mm} 
\\
HQ \hspace{-4mm} &
Mosaiced \hspace{-4mm} &
IRCNN~\cite{zhang2017learning} \hspace{-4mm} &
RNAN~\cite{zhang2019residual} \hspace{-4mm} &
PANet 
\\
\end{tabular}
\end{adjustbox}
%\vspace{2mm}
%\\

\end{tabular}
\vspace{-2mm}
\caption{Visual image demosaicing results}
\label{fig:result_Demosaic_N1_main}
\end{figure}

The visual results are shown in Fig.~\ref{fig:result_Demosaic_N1_main}. While IRCNN and RNAN can effectively reduce mosaic corruption to some degree, their results still contain evident artifacts. By relying on pyramid attention, the proposed PANet removed most of blocking artifacts and restores the more accurate color, as compared to IRCNN and RNAN.
\begin{table}[t] \label{tab:car}
%\scriptsize
%\footnotesize
%\small
%\normalsize
\center
\begin{center}
\caption{Quantitative evaluation of state-of-the-art approaches on compression artifacts reduction. Best results are \textbf{highlighted}}
\label{tab:results_psnr_ssim_car_y}
\vspace{-4mm}
\resizebox{1.0\columnwidth}{!}{
%\begin{tabular*}{134mm}{|c|c|c@{\extracolsep{-0.99mm}}|c|c|c|c|c|c|c|c|c|c|c|c|c|}
\begin{tabular}{|l|c|c|c|c|c|c|c|c|c|c|c|c|c|c|c|c|}
\hline
\multirow{2}{*}{Dataset} & \multirow{2}{*}{$q$} &  \multicolumn{2}{c|}{JPEG} &  \multicolumn{2}{c|}{SA-DCT} &  \multicolumn{2}{c|}{ARCNN} &  \multicolumn{2}{c|}{TNRD} &  \multicolumn{2}{c|}{DnCNN} &  \multicolumn{2}{c|}{RNAN} &    
\multicolumn{2}{c|}{PANet} 
\\
%\hline
\cline{3-16}
&  & PSNR & SSIM & PSNR & SSIM & PSNR & SSIM & PSNR & SSIM & PSNR & SSIM & PSNR & SSIM  &PSNR &SSIM
\\
\hline
\hline
\multirow{4}{*}{LIVE1} & 10 
& 27.77
 & 0.7905
  & 28.65
   & 0.8093
    & 28.98
     & 0.8217
      & 29.15
       & 0.8111
        & {29.19}
         & {0.8123}
          & \underline{29.63}
           & \underline{0.8239}
            &\textbf{29.69}
            &\textbf{0.8250}

\\
& 20 
& 30.07
 & 0.8683
  & 30.81
   & 0.8781
    & 31.29
     & 0.8871
      & 31.46
       & 0.8769
        & {31.59}
         & {0.8802}
          & \underline{32.03}
           & \underline{0.8877}
            &\textbf{32.10}
             &\textbf{0.8885}

\\
& 30 
& 31.41
 & 0.9000
  & 32.08
   & 0.9078
    & 32.69
     & 0.9166
      & 32.84
       & 0.9059
        & {32.98}
         & {0.9090}
          & \underline{33.45}
           & \underline{0.9149}
            &\textbf{33.55}
            &\textbf{0.9157}

\\
& 40 
& 32.35
 & 0.9173
  & 32.99
   & 0.9240
    & 33.63
     & 0.9306
      & N/A
       & N/A
        & {33.96}
         & {0.9247}
          & \underline{34.47}
           & \underline{0.9299}
            & \textbf{34.55}
            &\textbf{0.9305}

\\
\hline 
\hline
\multirow{4}{*}{Classic5} & 10 
& 27.82
 & 0.7800
  & 28.88
   & 0.8071
    & 29.04
     & 0.8111
      & 29.28
       & 0.7992
        & {29.40}
         & {0.8026}
          & \underline{29.96}
           & \underline{0.8178}
            & \textbf{30.03}
            &\textbf{0.8195}

\\
& 20 
& 30.12
 & 0.8541
  & 30.92
   & 0.8663
    & 31.16
     & 0.8694
      & 31.47
       & 0.8576
        & {31.63}
         & {0.8610}
          & \underline{32.11}
           & \underline{0.8693}
            &\textbf{32.36}
            &\textbf{0.8712}

\\
& 30 
& 31.48
 & 0.8844
  & 32.14
   & 0.8914
    & 32.52
     & 0.8967
      & 32.78
       & 0.8837
        & {32.91}
         & {0.8861}
          & \underline{33.38}
           & \underline{0.8924}
           &\textbf{33.53}
           &\textbf{0.8939}

\\
& 40 
& 32.43
 & 0.9011
  & 33.00
   & 0.9055
    & 33.34
     & 0.9101
      & N/A
       & N/A
        & {33.77}
         & {0.9003}
          & \underline{34.27}
           & \underline{0.9061}
            &\textbf{34.38}
             &\textbf{0.9068}

\\
\hline          
\end{tabular}
}
\end{center}

\end{table}
\subsection{Image Compression Artifacts Reduction}
For image compression artifacts reduction (CAR), we compare our method with 5 most recent approaches: SA-DCT~\cite{foi2007pointwise}, ARCNN~\cite{dong2015compression}, TNRD~\cite{chen2017trainable}, DnCNN~\cite{zhang2017beyond}, and RNAN~\cite{zhang2019residual}. We present results on LIVE1~\cite{sheikh2005live} and Classic5~\cite{foi2007pointwise}, following the same settings in RNAN. To obtain the low-quality compressed images, we follow the standard JPEG compression process and use Matlab JPEG encoder with quality $q = 10, 20, 30, 40$.  For fair comparison, the results are only evaluated on Y channel in YCbCr Space.

\begin{figure}[tpb]
%\newlength-4mm
%\setlength{-4mm}{-0.4cm}
\scriptsize
\centering
\begin{tabular}{cc}
% % one row
%\hspace{-0.2cm}
\begin{adjustbox}{valign=t}
\begin{tabular}{c}
\includegraphics[width=0.1535\textwidth,height=0.134\textwidth]{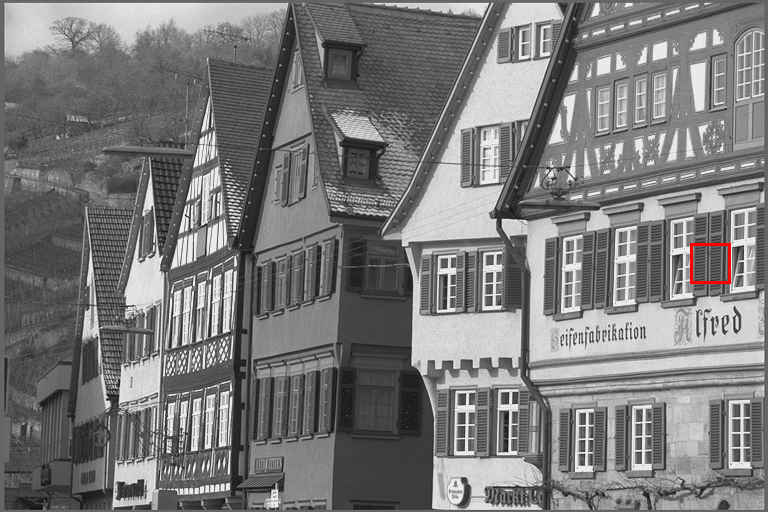}
\\
Buildings
\end{tabular}
\end{adjustbox}
\hspace{-0.23cm}
\begin{adjustbox}{valign=t}
\begin{tabular}{cccccc}
\includegraphics[width=0.134\textwidth]{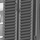} \hspace{-1.5mm} &
\includegraphics[width=0.134\textwidth]{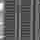} \hspace{-1.5mm} &
\includegraphics[width=0.134\textwidth]{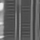} \hspace{-1.5mm} &
\includegraphics[width=0.134\textwidth]{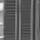} \hspace{-1.5mm} &
\includegraphics[width=0.134\textwidth]{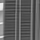} \hspace{-1.5mm} &
\includegraphics[width=0.134\textwidth]{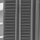} \hspace{-1.5mm}
\\
HQ \hspace{-1.5mm} &
JPEG \hspace{-1.5mm} &
ARCNN~\cite{dong2015compression} \hspace{-1.5mm} &
DnCNN~\cite{zhang2017beyond} \hspace{-1.5mm} &
RNAN~\cite{zhang2019residual} \hspace{-1.5mm} &
PANet
\\
\end{tabular}
\end{adjustbox}
\vspace{2mm}
\\

% % one row
\hspace{-0.15cm}
\begin{adjustbox}{valign=t}
\begin{tabular}{c}
\includegraphics[width=0.1535\textwidth,height=0.134\textwidth]{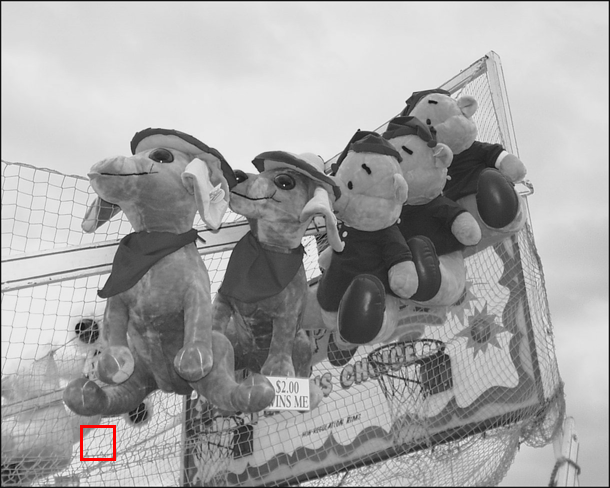}
\\
Carnivaldolls
\end{tabular}
\end{adjustbox}
\hspace{-0.23cm}
\begin{adjustbox}{valign=t}
\begin{tabular}{cccccc}
\includegraphics[width=0.134\textwidth]{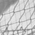} \hspace{-1.5mm} &
\includegraphics[width=0.134\textwidth]{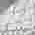} \hspace{-1.5mm} &
\includegraphics[width=0.134\textwidth]{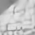} \hspace{-1.5mm} &
\includegraphics[width=0.134\textwidth]{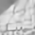} \hspace{-1.5mm} &
\includegraphics[width=0.134\textwidth]{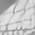} \hspace{-1.5mm} &
\includegraphics[width=0.134\textwidth]{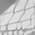} \hspace{-1.5mm}
\\
HQ \hspace{-1.5mm} &
JPEG \hspace{-1.5mm} &
ARCNN~\cite{dong2015compression} \hspace{-1.5mm} &
DnCNN~\cite{zhang2017beyond} \hspace{-1.5mm} &
RNAN~\cite{zhang2019residual} \hspace{-1.5mm} &
PANet
\\
\end{tabular}
\end{adjustbox}
%\vspace{2mm}
%\\

\end{tabular}
\vspace{-2mm}
\caption{Visual comparison for image CAR with JPEG quality $q$ = 10}
\label{fig:result_CAY_Y_Q10_main}
\end{figure}

The quantitative evaluation are reported in Table \ref{tab:results_psnr_ssim_car_y}. By incorporating pyramid attention, PANet obtains best results on both LIVE1 and Classic5 with all quality levels. We further present visual comparisons on the most challenging quality level $q=10$ in Fig.~\ref{fig:result_CAY_Y_Q10_main}. One can see that the proposed approach successfully reduced compression artifacts and recovered the most image details. This is mainly because our PANet captures non-local relationships in a multi-scale way, helping to reconstruct more faithful details. 
\vspace{-2mm}

% % CAR_Y
\begin{table}[h]\setlength{\tabcolsep}{5pt}
\footnotesize
%\small
\centering
\caption{Model size comparison}
\vspace{-3mm}
\label{tab:model_size}\resizebox{1.0\columnwidth}{!}{
\begin{tabular}{|c|c|c|c|c|c|c|c|}
\hline
Methods & RED  & DnCNN & MemNet &RNAN(1LB1NL) & RNAN & PANet-S & PANet\\
\hline
\hline
Parameters &4131K &672K & 677K &1494K& 7409K &\textbf{665K}& 5957K \\
\hline
PSNR (dB) &26.40 &28.16 & 26.53 &28.36& 29.15 & 28.80 &\textbf{29.47}\\
\hline
\end{tabular}}
\vspace{-2mm}
\end{table}

\vspace{-3mm}
\subsection{Model Size Analyses}
We report our model size and compare it with other advanced image denoising approaches in Table~\ref{tab:model_size}. To compare with light weight models, we also bulid a small PANet-S with only 8 residual blocks. One can see that PANet achieves the best performance with a lighter much simpler architecture, as compared to the prior state-of-the-art approach RNAN. Similarly, PANet-S significantly outperforms other light weight models using only less than 50\% parameters of RNAN (1LB+1NLB). Such observations indicate the great advantages brought by our pyramid attention module. In practice, our proposed pyramid attention module can be inserted in related networks.
\begin{table*}[t]\setlength{\tabcolsep}{7pt}
%\scriptsize
%\footnotesize
\small
%\normalsize
\center
\begin{center}
\caption{Quantitative results on SR benchmark datasets}
\vspace{-7mm}
\label{tab:results_psnr_ssim_x2348}
\resizebox{1.0\columnwidth}{!}{
\begin{tabular}{|l|c|c|c|c|c|c|c|c|c|c|c|}
\hline
\multirow{2}{*}{Method} & \multirow{2}{*}{Scale} &  \multicolumn{2}{c|}{Set5} &  \multicolumn{2}{c|}{Set14} &  \multicolumn{2}{c|}{B100} &  \multicolumn{2}{c|}{Urban100} &  \multicolumn{2}{c|}{Manga109}  
\\
%\hline
\cline{3-12}
&  & PSNR & SSIM & PSNR & SSIM & PSNR & SSIM & PSNR & SSIM & PSNR & SSIM 
\\
\hline
\hline

LapSRN~\cite{lai2017deep} & $\times$2 
& 37.52
 & 0.9591
  & 33.08
   & 0.9130
    & 31.08
     & 0.8950
      & 30.41
       & 0.9101
        & 37.27
         & 0.9740
                   
\\
MemNet~\cite{tai2017memnet} & $\times$2 
& 37.78
 & 0.9597
  & 33.28
   & 0.9142
    & 32.08
     & 0.8978
      & 31.31
       & 0.9195
        & 37.72
         & 0.9740

\\
SRMDNF~\cite{zhang2018learning} & $\times$2 
& 37.79
 & 0.9601
  & 33.32
   & 0.9159
    & 32.05
     & 0.8985
      & 31.33
       & 0.9204
        & 38.07
         & 0.9761
                   
\\
DBPN~\cite{haris2018deep} & $\times$2 
& 38.09
 & 0.9600
  & 33.85
   & 0.9190
    & 32.27
     & 0.9000
      & 32.55
       & 0.9324
        & 38.89
         & 0.9775        
\\
RDN~\cite{zhang2018residual} & $\times$2 
& 38.24
 & 0.9614
  & 34.01
   & 0.9212
    & 32.34
     & 0.9017
      & 32.89
       & 0.9353
        & 39.18
         & 0.9780
         
\\

RCAN~\cite{zhang2018image} & $\times$2 
& {38.27}
 & {0.9614}
  & \underline{34.12}
   & \underline{0.9216}
    & {32.41}
     &{0.9027}
      & \underline{33.34}
       & \underline{0.9384}
        & \textbf{39.44}
         & {0.9786}
\\       
NLRN~\cite{liu2018non}& $\times$2 
& {38.00}
 & {0.9603}
  & {33.46}
   & {0.9159}
    & {32.19}
     & {0.8992}
      & {31.81}
       & {0.9249}
        & {--}
         & {--}
\\
SRFBN~\cite{li2019feedback}& $\times$2 
& {38.11}
 & {0.9609}
  & {33.82}
   & {0.9196}
    & {32.29}
     & {0.9010}
      & {32.62}
       & {0.9328}
        & {39.08}
         &{0.9779}
\\
 OISR~\cite{he2019ode} & $\times$2 
& {38.21}
 & {0.9612}
  & {33.94}
   & {0.9206}
    & {32.36}
     & {0.9019}
      & {33.03}
       & {0.9365}
        &--
         & --      
\\
SAN~\cite{dai2019second} & $\times$2 
& \underline{38.31}
 & \textbf{0.9620}
  & {34.07}
   & {0.9213}
    & \underline{32.42}
     & \textbf{0.9028}
      & \underline{33.10}
       & \underline{0.9370}
        & {39.32}
         & \textbf{0.9792}
\\
\hline
EDSR~\cite{lim2017enhanced} & $\times$2 
& 38.11
 & 0.9602
  & 33.92
   & 0.9195
    & 32.32
     & 0.9013
      & 32.93
       & 0.9351
        & 39.10
         & 0.9773
         \\
PA-EDSR (ours) & $\times$2 
& \textbf{38.33}
 & \underline{0.9617}
  & \textbf{34.22}
   & \textbf{0.9224}
    & \textbf{32.42}
     & \textbf{0.9027}
      & \textbf{33.38}
       & \textbf{0.9392}
        & \underline{39.37}
         & \underline{0.9782}
\\

\hline                 
\hline

LapSRN~\cite{lai2017deep} & $\times$3 
& 33.82
 & 0.9227
  & 29.87
   & 0.8320
    & 28.82
     & 0.7980
      & 27.07
       & 0.8280
        & 32.21
         & 0.9350
                   
\\
MemNet~\cite{tai2017memnet} & $\times$3 
& 34.09
 & 0.9248
  & 30.00
   & 0.8350
    & 28.96
     & 0.8001
      & 27.56
       & 0.8376
        & 32.51
         & 0.9369

\\
SRMDNF~\cite{zhang2018learning} & $\times$3 
& 34.12
 & 0.9254
  & 30.04
   & 0.8382
    & 28.97
     & 0.8025
      & 27.57
       & 0.8398
        & 33.00
         & 0.9403
                   
\\
RDN~\cite{zhang2018residual} & $\times$3 
& 34.71
 & 0.9296
  & 30.57
   & 0.8468
    & 29.26
     & 0.8093
      & 28.80
       & 0.8653
        & 34.13
         & 0.9484
         
\\
RCAN~\cite{zhang2018image}& $\times$3 
& {34.74}
 &{0.9299}
  & {30.65}
   & \underline{0.8482}
    & {29.32}
     & {0.8111}
      & \underline{29.09}
       &\underline{0.8702}
        & \underline{34.44}
         &\underline{0.9499}
         
\\
NLRN~\cite{liu2018non}& $\times$3 
& {34.27}
 &{0.9266}
  & {30.16}
   &{0.8374}
    & {29.06}
     & {0.8026}
      & {27.93}
       & {0.8453}
        & {-}
         & {-}
\\
SRFBN~\cite{li2019feedback}& $\times$3 
& {34.70}
 &{0.9292}
  & {30.51}
   &{0.8461}
    & {29.24}
     & {0.8084}
      & {28.73}
       & {0.8641}
        & {34.18}
         & {0.9481}
\\
OISR~\cite{he2019ode}& $\times$3 
& {34.72}
 &{0.9297}
  & {30.57}
   &{0.8470}
    & {29.29}
     & {0.8103}
      & {28.95}
       & {0.8680}
        & {-}
         & {-}
\\
SAN~\cite{dai2019second} & $\times$3 
& \underline{34.75}
 &\underline{0.9300}
  & {30.59}
   &{0.8476}
    &\underline{29.33}
     & \underline{0.8112}
      & {28.93}
       & {0.8671}
        & {34.30}
         & {0.9494}
        
\\
\hline
EDSR~\cite{lim2017enhanced} & $\times$3 
& 34.65
 & 0.9280
  & 30.52
   & 0.8462
    & 29.25
     & 0.8093
      & 28.80
       & 0.8653
        & 34.17
         & 0.9476
\\
PA-EDSR (ours) & $\times$3
& \textbf{34.84}
 & \textbf{0.9306}
  & \textbf{30.71}
   & \textbf{0.8488}
    & \textbf{29.33}
     & \textbf{0.8119}
      & \textbf{29.24}
       & \textbf{0.8736}
        & \textbf{34.46}
         & \textbf{0.9505}
\\
\hline
\hline
LapSRN~\cite{lai2017deep} & $\times$4 
& 31.54
 & 0.8850
  & 28.19
   & 0.7720
    & 27.32
     & 0.7270
      & 25.21
       & 0.7560
        & 29.09
         & 0.8900
                   
\\
MemNet~\cite{tai2017memnet} & $\times$4 
& 31.74
 & 0.8893
  & 28.26
   & 0.7723
    & 27.40
     & 0.7281
      & 25.50
       & 0.7630
        & 29.42
         & 0.8942
                   
\\
SRMDNF~\cite{zhang2018learning} & $\times$4 
& 31.96
 & 0.8925
  & 28.35
   & 0.7787
    & 27.49
     & 0.7337
      & 25.68
       & 0.7731
        & 30.09
         & 0.9024
                   
\\
DBPN~\cite{haris2018deep} & $\times$4 
& 32.47
 & 0.8980
  & 28.82
   & 0.7860
    & 27.72
     & 0.7400
      & 26.38
       & 0.7946
        & 30.91
         & 0.9137
         
\\
RDN~\cite{zhang2018residual} & $\times$4 
& 32.47
 & 0.8990
  & 28.81
   & 0.7871
    & 27.72
     & 0.7419
      & 26.61
       & 0.8028
        & 31.00
         & 0.9151
         
\\
RCAN~\cite{zhang2018image}& $\times$4 
& {32.63}
 & {0.9002}
  & {28.87}
   &\underline{0.7889}
    & \underline{27.77}
     & \underline{0.7436}
      &\underline{26.82}
       & \underline{0.8087}
        &\underline{31.22}
         & \underline{0.9173}

\\
NLRN~\cite{liu2018non}& $\times$4 
& {31.92}
 & {0.8916}
  & {28.36}
   & {0.7745}
    & {27.48}
     & {0.7306}
      & {25.79}
       & {0.7729}
        & {-}
         & {-}
\\
SRFBN~\cite{li2019feedback} & $\times$4 
& {32.47}
 & {0.8983}
  & {28.81}
   & {0.7868}
    & {27.72}
     & {0.7409}
      & {26.60}
       & {0.8015}
        & {31.15}
         & {0.9160}
\\
OISR~\cite{he2019ode} & $\times$4 
&{32.53}
 &{0.8992}
  &{28.86}
   & {0.7878}
    &{27.75}
     & {0.7428}
      & {26.79}
       & {0.8068}
        & {-}
         & {-}
\\
SAN~\cite{dai2019second} & $\times$4 
& \underline{32.64}
 &\underline{0.9003}
  &\textbf{28.92}
   &\underline{0.7888}
    &\textbf{27.78}
     & \underline{0.7436}
      & {26.79}
       & {0.8068}
        & {31.18}
         & {0.9169}
\\
\hline
EDSR~\cite{lim2017enhanced} & $\times$4 
& 32.46
 & 0.8968
  & 28.80
   & 0.7876
    & 27.71
     & 0.7420
      & 26.64
       & 0.8033
        & 31.02
         & 0.9148
         \\
PA-EDSR (ours)  & $\times$4 
& \textbf{32.65}
 & \textbf{0.9006}
  & \underline{28.87}
   & \textbf{0.7891}
    & {27.76}
     & \textbf{0.7445}
      & \textbf{27.01}
       & \textbf{0.8140}
        & \textbf{31.29}
         & \textbf{0.9194}

\\
\hline             
\end{tabular}
}
\end{center}
\end{table*}

\begin{figure*}[h]
%\tiny
%\small
	\newlength\fsdttwofigBD
	\setlength{\fsdttwofigBD}{-0.9mm}
		\caption{
		Visual comparison for $4\times$ SR on Urban100 dataset}
		\label{figure:sr}
	\scriptsize
	\centering
	\begin{tabular}{cc}
	%\tiny
	%\scriptsize
	%\footnotesize
	%\small
		%\hspace{-0.4cm}
		\begin{adjustbox}{valign=t}
		%\tiny
			\begin{tabular}{c}
				\includegraphics[height=0.25\textwidth, width=0.229\textwidth]{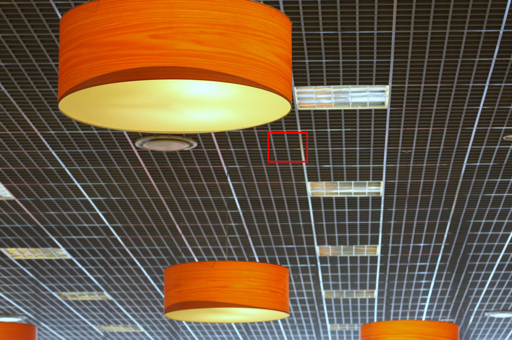}
				\\
				 Urban100 ($4\times$):
				\\
				img\_044
				%\textsc{B100}: img_092
				
			\end{tabular}
		\end{adjustbox}
		\hspace{-2.3mm}
		\begin{adjustbox}{valign=t}
		%\tiny
			\begin{tabular}{cccccc}
				\includegraphics[width=\widthscalefive \textwidth]{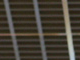} \hspace{\fsdttwofigBD} &
				\includegraphics[width=\widthscalefive \textwidth]{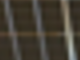} \hspace{\fsdttwofigBD} &
				\includegraphics[width=\widthscalefive \textwidth]{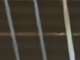} \hspace{\fsdttwofigBD} &
				\includegraphics[width=\widthscalefive \textwidth]{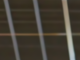} \hspace{\fsdttwofigBD} &
				\includegraphics[width=\widthscalefive \textwidth]{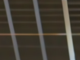} 
				\\
				HR \hspace{\fsdttwofigBD} &
				Bicubic \hspace{\fsdttwofigBD} &
				LapSRN~\cite{lai2017deep} \hspace{\fsdttwofigBD} &
				EDSR~\cite{lim2017enhanced} \hspace{\fsdttwofigBD} &
				DBPN~\cite{haris2018deep}
				\\
				\includegraphics[width=\widthscalefive \textwidth]{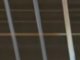} \hspace{\fsdttwofigBD} &
				\includegraphics[width=\widthscalefive \textwidth]{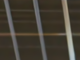} \hspace{\fsdttwofigBD} &
				\includegraphics[width=\widthscalefive \textwidth]{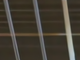} \hspace{\fsdttwofigBD} &
				\includegraphics[width=\widthscalefive \textwidth]{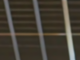} \hspace{\fsdttwofigBD} &
				\includegraphics[width=\widthscalefive \textwidth]{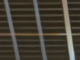}  
				\\ 
				OISR~\cite{he2019ode} \hspace{\fsdttwofigBD} &
				RDN~\cite{zhang2018residual} \hspace{\fsdttwofigBD} &
				RCAN~\cite{zhang2018image} \hspace{\fsdttwofigBD} &
				SAN~\cite{dai2019second}  \hspace{\fsdttwofigBD} &
				Ours
			 \hspace{\fsdttwofigBD} 
				\\
			\end{tabular}
		\end{adjustbox}
		\vspace{0.5mm}
		\\

		\begin{adjustbox}{valign=t}
		%\tiny
			\begin{tabular}{c}
				\includegraphics[width=0.229\textwidth, height =0.25\textwidth]{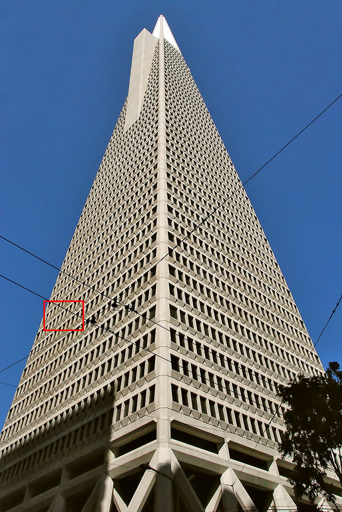}
				\\
				 Urban100 ($4\times$):
				\\
				img\_048
				%\textsc{B100}: img_092
				
			\end{tabular}
		\end{adjustbox}
		\hspace{-2.3mm}
		\begin{adjustbox}{valign=t}
		%\tiny
			\begin{tabular}{cccccc}
				\includegraphics[width=\widthscalefive \textwidth]{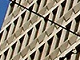} \hspace{\fsdttwofigBD} &
				\includegraphics[width=\widthscalefive \textwidth]{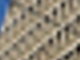} \hspace{\fsdttwofigBD} &
				\includegraphics[width=\widthscalefive \textwidth]{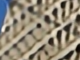} \hspace{\fsdttwofigBD} &
				\includegraphics[width=\widthscalefive \textwidth]{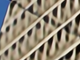} \hspace{\fsdttwofigBD} &
				\includegraphics[width=\widthscalefive \textwidth]{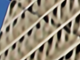} 
				\\
				HR \hspace{\fsdttwofigBD} &
				Bicubic \hspace{\fsdttwofigBD} &
				LapSRN~\cite{lai2017deep} \hspace{\fsdttwofigBD} &
				EDSR~\cite{lim2017enhanced} \hspace{\fsdttwofigBD} &
				DBPN~\cite{haris2018deep}
				\\
				\includegraphics[width=\widthscalefive \textwidth]{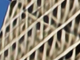} \hspace{\fsdttwofigBD} &
				\includegraphics[width=\widthscalefive \textwidth]{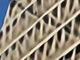} \hspace{\fsdttwofigBD} &
				\includegraphics[width=\widthscalefive \textwidth]{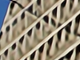} \hspace{\fsdttwofigBD} &
				\includegraphics[width=\widthscalefive \textwidth]{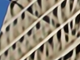} \hspace{\fsdttwofigBD} &
				\includegraphics[width=\widthscalefive \textwidth]{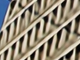}  
				\\ 
				
			    OISR~\cite{he2019ode} \hspace{\fsdttwofigBD} &
				RDN~\cite{zhang2018residual} \hspace{\fsdttwofigBD} &
				RCAN~\cite{zhang2018image} \hspace{\fsdttwofigBD} &
				SAN~\cite{dai2019second}  \hspace{\fsdttwofigBD} &
				Ours
				\\
			\end{tabular}
		\end{adjustbox}
		\vspace{0.5mm}
		\\
		%\hspace{-0.4cm}		
		\begin{adjustbox}{valign=t}
		%\tiny
			\begin{tabular}{c}
				\includegraphics[width=0.229\textwidth, height=0.25\textwidth]{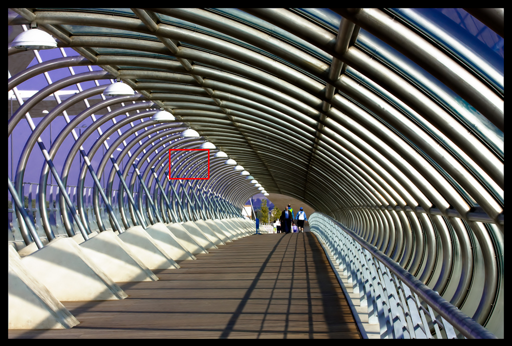}
				\\
				Urban100 ($4\times$):
				\\
				%\textsc{Urban100}: img\_004
				img\_058
			\end{tabular}
		\end{adjustbox}
		\hspace{-2.3mm}
		\begin{adjustbox}{valign=t}
		%\tiny
			\begin{tabular}{cccccc}
				\includegraphics[width=\widthscalefive \textwidth]{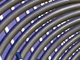} \hspace{\fsdttwofigBD} &
				\includegraphics[width=\widthscalefive \textwidth]{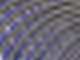} \hspace{\fsdttwofigBD} &
				\includegraphics[width=\widthscalefive \textwidth]{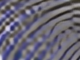} \hspace{\fsdttwofigBD} &
				\includegraphics[width=\widthscalefive \textwidth]{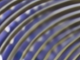} \hspace{\fsdttwofigBD} &
				\includegraphics[width=\widthscalefive \textwidth]{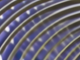} 
				\\
				HR \hspace{\fsdttwofigBD} &
				Bicubic \hspace{\fsdttwofigBD} &
				LapSRN~\cite{lai2017deep} \hspace{\fsdttwofigBD} &
				EDSR~\cite{lim2017enhanced} \hspace{\fsdttwofigBD} &
				DBPN~\cite{haris2018deep}
				\\
				\includegraphics[width=\widthscalefive \textwidth]{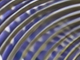} \hspace{\fsdttwofigBD} &
				\includegraphics[width=\widthscalefive \textwidth]{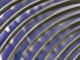} \hspace{\fsdttwofigBD} &
				\includegraphics[width=\widthscalefive \textwidth]{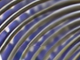} \hspace{\fsdttwofigBD} &
				\includegraphics[width=\widthscalefive \textwidth]{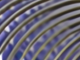} \hspace{\fsdttwofigBD} &
				\includegraphics[width=\widthscalefive \textwidth]{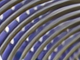}  
				\\ 
				OISR~\cite{he2019ode} \hspace{\fsdttwofigBD} &
				RDN~\cite{zhang2018residual} \hspace{\fsdttwofigBD} &
				RCAN~\cite{zhang2018image} \hspace{\fsdttwofigBD} &
				SAN~\cite{dai2019second}  \hspace{\fsdttwofigBD} &
				Ours
				\\
				
				\\
			\end{tabular}
		\end{adjustbox}
	\end{tabular}
\vspace{-3mm}
\end{figure*}
\subsection{Image Super Resolution}
To further demonstrate the generality of pyramid attention, we present additional image super resolution experiments. Similar to previous settings, we build our network on top of the widely used plain ResNet based EDSR \cite{lim2017enhanced}, where a single pyramid attention block is inserted after 16th residual block. We compare it with 11 state-of-the-art approches: LapSRN \cite{lai2017deep}, MemNet \cite{tai2017memnet}, SRMDNF \cite{zhang2017learning}, DBPN \cite{haris2018deep}, RDN \cite{zhang2018residual}, RCAN \cite{zhang2018image}, NLRN \cite{liu2018non}, SRFBN \cite{li2019feedback}, OISR \cite{he2019ode}, and SAN \cite{dai2019second}. 

We report experiment results in Table \ref{tab:results_psnr_ssim_x2348}. Without any architectural engineering, our simple PA-EDSR achieves best performance on almost all benchmarks and scales. The significant improvements over EDSR demonstrate the effectiveness of the proposed attention operation. One may notice that the improvements are not limit to Urban100, where images contain apparent structural recurrences. We also observed considerable performance gain on natural image datasets: Set5, Set14 and B100. This is accorded with previous observation that cross-scale self-recurrences is a common property for natural images \cite{glasner2009super}. The proposed operation is a generic block and therefore can be also integrated into other SR networks to further boost SR performance.

Visual results are shown in Fig. \ref{figure:sr}. Our PA-EDSR reconstructs the most accurate image details, leading to better visual pleasing results.

\subsection{Visualization of Attention Map.}
\begin{figure}[h]
%\newlength-4mm
%\setlength{-4mm}{-0.4cm}
\scriptsize
\centering
\hspace{-1mm}
\begin{tabular}{c}
% % one row
\begin{adjustbox}{valign=t}
\begin{tabular}{cccccc}
\includegraphics[width=0.16\textwidth]{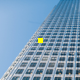} &
\includegraphics[width=0.16\textwidth]{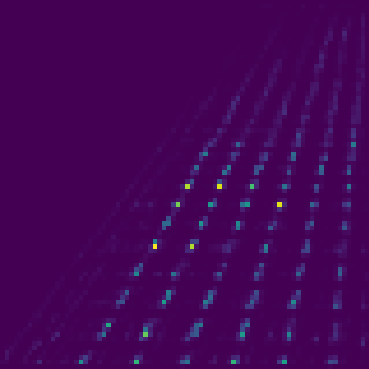} &
\includegraphics[width=0.16\textwidth]{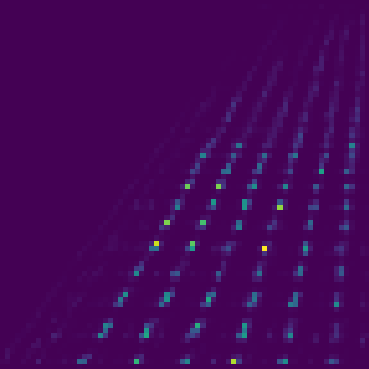}&
\includegraphics[width=0.16\textwidth]{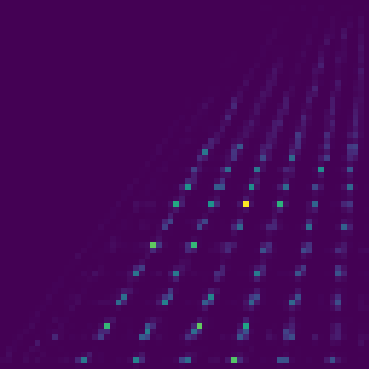} &
\includegraphics[width=0.16\textwidth]{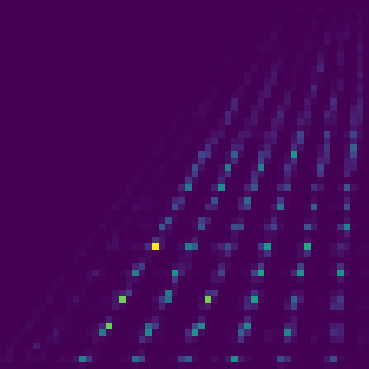} &
\includegraphics[width=0.16\textwidth]{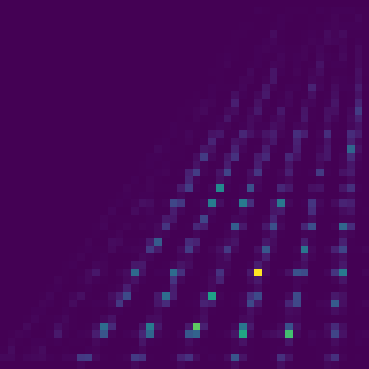} 
\\
\includegraphics[width=0.16\textwidth]{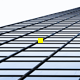}  &
\includegraphics[width=0.16\textwidth]{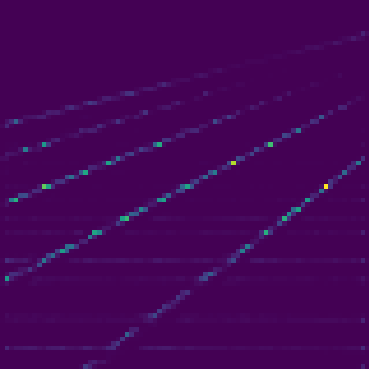} &
\includegraphics[width=0.16\textwidth]{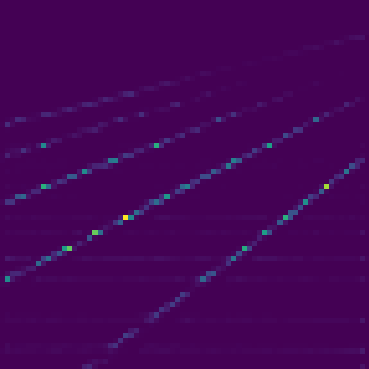}  &
\includegraphics[width=0.16\textwidth]{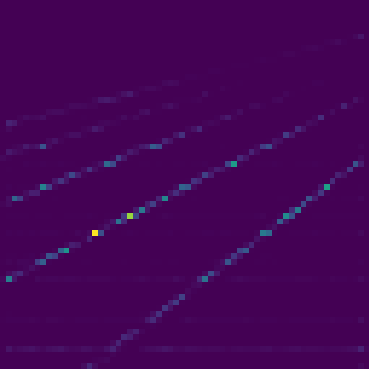}  &
\includegraphics[width=0.16\textwidth]{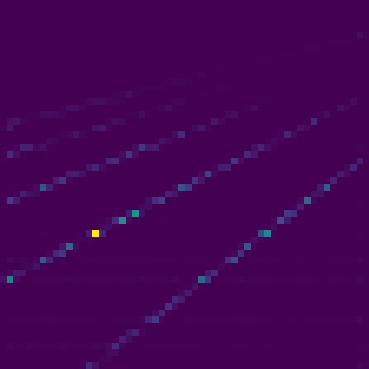} &  
\includegraphics[width=0.16\textwidth]{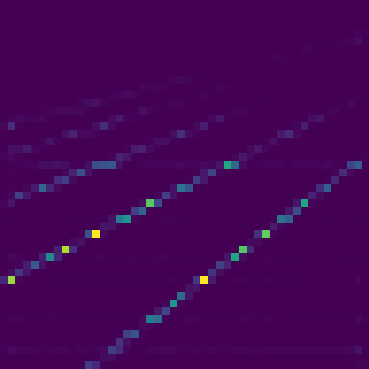} 
\\ 
HQ  &
Level1 &
Level2 &
Level3 &
Level4 &
Level5
\\
\end{tabular}
\end{adjustbox}
\end{tabular}
%\vspace{2mm}
\vspace{-2mm}
\caption{Visualization of correlation maps of pyramid attention. Maps are rescaled to same size for visualization purpose. Brighter color indicates higher engagement. One can see that the attention focuses on different locations at each scale, indicating the module is able to exploit multi-scale recurrences to improve restoration}
\label{fig:map}
\vspace{-4mm}
\end{figure}

To fully demonstrate that our pyramid attention captures multi-scale correlations, we visualize its attention map in Fig. \ref{fig:map}. For illustration purpose, the selected images contain abundant self-exemplars at different locations and scales. 

From Fig. \ref{fig:map}, we find the attention maps follow distinct distributions over scales, demonstrating that our attention is able to focus on informative regions at multiple scales. It is interesting to point out, as level increases, the most engaged patches move downwards. This is in line with that larger patterns, such as window, appear at bottom in selected images. By capturing multi-scale correlations, the network managed to utilize these informative patches to improve restoration.

\begin{table}[t]
%\footnotesize
%\small
\centering
\caption{Effects of pyramid attention on Urban100 ($\sigma=30$)}
\vspace{-2mm}
\label{tab:PA}
\begin{tabular}{|l|c|c|c|}
\hline
& baseline & N-L attention  & Pyramid attention\\
\hline
\hline
PSNR &30.86 &31.14 & \textbf{31.29} \\
\hline
\end{tabular}
\vspace{-3mm}
\end{table}

\begin{table}[t]\setlength{\tabcolsep}{5pt}
%\footnotesize
%\small
\centering
%\vspace{2mm}
\caption{Comparison between pixel-wise matching and block-wise matching on Urban100 ($\sigma=30$)}
\label{tab:matching}
\vspace{-2mm}
\begin{tabular}{|l|c|c|c|}
\hline
& baseline & pixel-wise  & block-wise\\
\hline
\hline
PSNR &30.86 &31.14 & \textbf{31.21} \\
\hline
\end{tabular}
\vspace{-3mm}
\end{table}

\begin{table}[hbp] 
%\footnotesize
%\small
    \centering
    \caption{Ablation study on pyramid levels}
    \label{tab:level}
    \vspace{-2mm}
    \begin{tabular}{|l|c||c|c|c|c|c|}
    \hline
          & baseline & 1-level & 2-level & 3-level & 4-level & 5-level \\
         \hline
         \hline
         PSNR &30.86 &31.21 & 31.23 & 31.25 & 31.28 & \textbf{31.29} \\
         \hline
    \end{tabular}
\vspace{-6mm}
\end{table}

\subsection{Ablation study}
\textbf{Pyramid Attention Module.}
To verify the effectiveness of pyramid attention, we conduct control experiments on image denosing tasks ($\sigma=30$). The baseline module is constructed by removing the attention block, resulting in a simple ResNet. We set the number of residual blocks $R=16$ in this experiment. In Table \ref{tab:PA}, \textit{baseline} achieves 30.86 dB on Urban100. To compare with classic non-local operations, we further construct a non-local baseline by replacing the pyramid attention with non-local attention. From the result in column 2, we can see that single-scale non-local operation is able to bring improvements. However, the best performance is achieved by using the proposed pyramid attention, with brings 0.43 dB over the baseline and 0.15 dB over classical non-local model. This indicates the proposed module can be served as a better alternative to model long-range dependency than current non-local operation. Such module can exploit informative corresponds exist at multiple image scales, which is of central importance for reconstructing more faithful images. 

\textbf{Matching: Pixel-wise v.s. Block-wise.} While classic non-local attentions compute pixel-wise feature correlation, we find block-wise matching yields much better restorations in practice. To demonstrate this, we compare conventional non-local operations with its patch-based alternative, where the patch size is set to be $3\times 3$. As shown in Table \ref{tab:matching}, when using block matching, the performance is improve from 31.14 dB to 31.21 dB. This is because block-matching involves extra similarity constraint on nearby pixels, thus can better distinguish highly relevant correspondences from noisy ones. These results demonstrate that small patches are indeed more robust descriptors for similarity measurements.

\begin{table}[t]
%\footnotesize
%\small
\centering
\caption{Results for models with pyramid attention inserted at different residual blocks on Urban100 ($\sigma=30$)}
\label{tab:table stage}
\vspace{-1mm}
\begin{tabular}{|l|c|c|c|c|c|c|c|c|}
\hline
        Pre &  & \cmark &  &  &\cmark &\cmark& &\cmark\\
    \cline{1-1}
         Mid & & &\cmark&\ &\cmark  & &\cmark&\cmark
         \\
         \cline{1-1}
         Post&  &  & &\cmark  & &\cmark &\cmark &\cmark \\
         \hline
         \hline
         PSNR &30.86 &31.07  &31.29 &31.18 &31.33 &31.33 &31.39 & \textbf{31.48} \\
         \hline
    \end{tabular}
\vspace{-5mm}
\end{table}

\textbf{Feature Pyramid Levels.}
As discussed above, the key difference between classic non-local operation and pyramid attention is that our module allows the network to utilize correspondences at multiple scales. Here we investigate the influences of pyramid levels. We conduct control experiments by gradually adding more levels to the feature pyramid. The final pyramid consists of 5 layers with scale factors $s = \{1,0.9 ,0.8, 0.7, 0.6\}$. As shown in Table \ref{tab:level}, when more layers are added, we observe constant performance gains. In column 6, the best performance is obtained when all levels are included. This is mainly because, as searching space is progressively expanded to more scales, the attention unit has higher possibilities to find a more informative correspondences that beyond original image scale. These results indicate that multi-scale relationship is essential for improving restoration.

\textbf{Position in Neural Networks.} Where should we add pyramid attention to the networks, in order to fully unleash its potential? Table \ref{tab:table stage} compares pyramid attentions inserted to different stages of a ResNet. Here we consider 3 typical positions: after the 1st residual block representing preprocessing, after the 8th residual block, which is the middle of the network, and after the last residual block representing post-processing. From the first 4 columns, we find that inserting our module at any stages bring evident improvements. The largest performance gain is achieved by inserting it at middle. Moreover, when multiple modules are combined, the restoration quality further boosts. The best result is achieved by including modules at all three positions.

%\input{./figs/attentionmap/fig_attention}
%#### conclusion ####

\vspace{-2mm}
\section{Conclusion}
\vspace{-2mm}
In this paper, we proposed a simple and generic pyramid attention for image restoration. The module generalizes classic self-attention to capture non-local relationships at multiple image scales. It is fully differentiable and can be used into any architectures. We demonstrate that modeling multi-scale correspondences brings significant improvements for the general image restoration tasks of image denosing, demosaicing, compression artifacts reduction and super resolution. On all tasks, a simple backbone with one pyramid attention achieves superior restoration accuracy over prior state-of-the-art approaches. We believe pyramid attention should be used as a common building block in future neural networks.

%\clearpage
% ---- Bibliography ----
%
% BibTeX users should specify bibliography style 'splncs04'.
% References will then be sorted and formatted in the correct style.
%
\bibliographystyle{splncs04}
\bibliography{denoise_conf}

\begin{thebibliography}{10}
\providecommand{\url}[1]{\texttt{#1}}
\providecommand{\urlprefix}{URL }
\providecommand{\doi}[1]{https://doi.org/#1}

\bibitem{bahat2017non}
Bahat, Y., Efrat, N., Irani, M.: Non-uniform blind deblurring by reblurring.
  In: Proceedings of the IEEE International Conference on Computer Vision. pp.
  3286--3294 (2017)

\bibitem{bahat2016blind}
Bahat, Y., Irani, M.: Blind dehazing using internal patch recurrence. In: 2016
  IEEE International Conference on Computational Photography (ICCP). pp.~1--9.
  IEEE (2016)

\bibitem{buades2005non}
Buades, A., Coll, B., Morel, J.M.: A non-local algorithm for image denoising.
  In: CVPR (2005)

\bibitem{buades2011non}
Buades, A., Coll, B., Morel, J.M.: Non-local means denoising. Image Processing
  On Line  \textbf{1},  208--212 (2011)

\bibitem{cao2019gcnet}
Cao, Y., Xu, J., Lin, S., Wei, F., Hu, H.: Gcnet: Non-local networks meet
  squeeze-excitation networks and beyond. In: Proceedings of the IEEE
  International Conference on Computer Vision Workshops. pp.~0--0 (2019)

\bibitem{chen2018learning}
Chen, C., Chen, Q., Xu, J., Koltun, V.: Learning to see in the dark. In:
  Proceedings of the IEEE Conference on Computer Vision and Pattern
  Recognition. pp. 3291--3300 (2018)

\bibitem{chen2017trainable}
Chen, Y., Pock, T.: Trainable nonlinear reaction diffusion: A flexible
  framework for fast and effective image restoration. TPAMI  (2017)

\bibitem{dabov2007color}
Dabov, K., Foi, A., Katkovnik, V., Egiazarian, K.: Color image denoising via
  sparse 3d collaborative filtering with grouping constraint in
  luminance-chrominance space. In: ICIP (2007)

\bibitem{dabov2007image}
Dabov, K., Foi, A., Katkovnik, V., Egiazarian, K.: Image denoising by sparse
  3-d transform-domain collaborative filtering. TIP  (2007)

\bibitem{dai2019second}
Dai, T., Cai, J., Zhang, Y., Xia, S.T., Zhang, L.: Second-order attention
  network for single image super-resolution. In: Proceedings of the IEEE
  Conference on Computer Vision and Pattern Recognition. pp. 11065--11074
  (2019)

\bibitem{dong2015compression}
Dong, C., Deng, Y., Change~Loy, C., Tang, X.: Compression artifacts reduction
  by a deep convolutional network. In: ICCV (2015)

\bibitem{fan2019scale}
Fan, Y., Yu, J., Liu, D., Huang, T.S.: Scale-wise convolution for image
  restoration. arXiv preprint arXiv:1912.09028  (2019)

\bibitem{foi2007pointwise}
Foi, A., Katkovnik, V., Egiazarian, K.: Pointwise shape-adaptive dct for
  high-quality denoising and deblocking of grayscale and color images. TIP
  (May 2007)

\bibitem{freedman2011image}
Freedman, G., Fattal, R.: Image and video upscaling from local self-examples.
  ACM Transactions on Graphics (TOG)  \textbf{30}(2),  1--11 (2011)

\bibitem{fu2019dual}
Fu, J., Liu, J., Tian, H., Li, Y., Bao, Y., Fang, Z., Lu, H.: Dual attention
  network for scene segmentation. In: Proceedings of the IEEE Conference on
  Computer Vision and Pattern Recognition. pp. 3146--3154 (2019)

\bibitem{glasner2009super}
Glasner, D., Bagon, S., Irani, M.: Super-resolution from a single image. In:
  2009 IEEE 12th international conference on computer vision. pp. 349--356.
  IEEE (2009)

\bibitem{haris2018deep}
Haris, M., Shakhnarovich, G., Ukita, N.: Deep back-projection networks for
  super-resolution. In: Proceedings of the IEEE conference on computer vision
  and pattern recognition. pp. 1664--1673 (2018)

\bibitem{he2010single}
He, K., Sun, J., Tang, X.: Single image haze removal using dark channel prior.
  IEEE transactions on pattern analysis and machine intelligence
  \textbf{33}(12),  2341--2353 (2010)

\bibitem{he2019ode}
He, X., Mo, Z., Wang, P., Liu, Y., Yang, M., Cheng, J.: Ode-inspired network
  design for single image super-resolution. In: Proceedings of the IEEE
  Conference on Computer Vision and Pattern Recognition. pp. 1732--1741 (2019)

\bibitem{huang2015single}
Huang, J.B., Singh, A., Ahuja, N.: Single image super-resolution from
  transformed self-exemplars. In: Proceedings of the IEEE conference on
  computer vision and pattern recognition. pp. 5197--5206 (2015)

\bibitem{kim2016accurate}
Kim, J., Kwon~Lee, J., Mu~Lee, K.: Accurate image super-resolution using very
  deep convolutional networks. In: CVPR (2016)

\bibitem{lai2017deep}
Lai, W.S., Huang, J.B., Ahuja, N., Yang, M.H.: Deep laplacian pyramid networks
  for fast and accurate super-resolution. In: CVPR (2017)

\bibitem{li2017aod}
Li, B., Peng, X., Wang, Z., Xu, J., Feng, D.: Aod-net: All-in-one dehazing
  network. In: Proceedings of the IEEE International Conference on Computer
  Vision. pp. 4770--4778 (2017)

\bibitem{li2019feedback}
Li, Z., Yang, J., Liu, Z., Yang, X., Jeon, G., Wu, W.: Feedback network for
  image super-resolution. In: Proceedings of the IEEE Conference on Computer
  Vision and Pattern Recognition. pp. 3867--3876 (2019)

\bibitem{lim2017enhanced}
Lim, B., Son, S., Kim, H., Nah, S., Lee, K.M.: Enhanced deep residual networks
  for single image super-resolution. In: CVPRW (2017)

\bibitem{liu2018non}
Liu, D., Wen, B., Fan, Y., Loy, C.C., Huang, T.S.: Non-local recurrent network
  for image restoration. In: NeurIPS (2018)

\bibitem{lotan2016needle}
Lotan, O., Irani, M.: Needle-match: Reliable patch matching under high
  uncertainty. In: Proceedings of the IEEE Conference on Computer Vision and
  Pattern Recognition. pp. 439--448 (2016)

\bibitem{mairal2009non}
Mairal, J., Bach, F., Ponce, J., Sapiro, G., Zisserman, A.: Non-local sparse
  models for image restoration. In: 2009 IEEE 12th international conference on
  computer vision. pp. 2272--2279. IEEE (2009)

\bibitem{mao2016image}
Mao, X., Shen, C., Yang, Y.B.: Image restoration using very deep convolutional
  encoder-decoder networks with symmetric skip connections. In: NeurIPS (2016)

\bibitem{martin2001database}
Martin, D., Fowlkes, C., Tal, D., Malik, J.: A database of human segmented
  natural images and its application to evaluating segmentation algorithms and
  measuring ecological statistics. In: ICCV (2001)

\bibitem{Mei2020image}
Mei, Y., Fan, Y., Zhou, Y., Huang, L., Huang, T.S., Shi, H.: Image
  super-resolution with cross-scale non-local attention and exhaustive
  self-exemplars mining. In: Proceedings of the IEEE Conference on Computer
  Vision and Pattern Recognition (CVPR) (2020)

\bibitem{michaeli2014blind}
Michaeli, T., Irani, M.: Blind deblurring using internal patch recurrence. In:
  European Conference on Computer Vision. pp. 783--798. Springer (2014)

\bibitem{paszke2017automatic}
Paszke, A., Gross, S., Chintala, S., Chanan, G., Yang, E., DeVito, Z., Lin, Z.,
  Desmaison, A., Antiga, L., Lerer, A.: Automatic differentiation in pytorch
  (2017)

\bibitem{roth2005fields}
Roth, S., Black, M.J.: Fields of experts: A framework for learning image
  priors. In: 2005 IEEE Computer Society Conference on Computer Vision and
  Pattern Recognition (CVPR'05). vol.~2, pp. 860--867. IEEE (2005)

\bibitem{sheikh2005live}
Sheikh, H.R., Wang, Z., Cormack, L., Bovik, A.C.: Live image quality assessment
  database release 2 (2005) (2005)

\bibitem{singh2014super}
Singh, A., Ahuja, N.: Super-resolution using sub-band self-similarity. In:
  Asian Conference on Computer Vision. pp. 552--568. Springer (2014)

\bibitem{tai2017memnet}
Tai, Y., Yang, J., Liu, X., Xu, C.: Memnet: A persistent memory network for
  image restoration. In: ICCV (2017)

\bibitem{timofte2017ntire}
Timofte, R., Agustsson, E., Van~Gool, L., Yang, M.H., Zhang, L., Lim, B., Son,
  S., Kim, H., Nah, S., Lee, K.M., et~al.: Ntire 2017 challenge on single image
  super-resolution: Methods and results. In: CVPRW (2017)

\bibitem{vincent2008extracting}
Vincent, P., Larochelle, H., Bengio, Y., Manzagol, P.A.: Extracting and
  composing robust features with denoising autoencoders. In: ICML (2008)

\bibitem{wang2018non}
Wang, X., Girshick, R., Gupta, A., He, K.: Non-local neural networks. In: CVPR
  (2018)

\bibitem{wang2004image}
Wang, Z., Bovik, A.C., Sheikh, H.R., Simoncelli, E.P.: Image quality
  assessment: from error visibility to structural similarity. TIP  (2004)

\bibitem{xia2019second}
Xia, B.N., Gong, Y., Zhang, Y., Poellabauer, C.: Second-order non-local
  attention networks for person re-identification. In: Proceedings of the IEEE
  International Conference on Computer Vision. pp. 3760--3769 (2019)

\bibitem{zhang2017beyond}
Zhang, K., Zuo, W., Chen, Y., Meng, D., Zhang, L.: Beyond a gaussian denoiser:
  Residual learning of deep cnn for image denoising. TIP  (2017)

\bibitem{zhang2017learning}
Zhang, K., Zuo, W., Gu, S., Zhang, L.: Learning deep cnn denoiser prior for
  image restoration. In: CVPR (2017)

\bibitem{zhang2017ffdnet}
Zhang, K., Zuo, W., Zhang, L.: Ffdnet: Toward a fast and flexible solution for
  cnn based image denoising. arXiv preprint arXiv:1710.04026  (2017)

\bibitem{zhang2018learning}
Zhang, K., Zuo, W., Zhang, L.: Learning a single convolutional super-resolution
  network for multiple degradations. In: Proceedings of the IEEE Conference on
  Computer Vision and Pattern Recognition. pp. 3262--3271 (2018)

\bibitem{zhang2018image}
Zhang, Y., Li, K., Li, K., Wang, L., Zhong, B., Fu, Y.: Image super-resolution
  using very deep residual channel attention networks. In: ECCV (2018)

\bibitem{zhang2019residual}
Zhang, Y., Li, K., Li, K., Zhong, B., Fu, Y.: Residual non-local attention
  networks for image restoration. In: ICLR (2019)

\bibitem{zhang2018residual}
Zhang, Y., Tian, Y., Kong, Y., Zhong, B., Fu, Y.: Residual dense network for
  image super-resolution. In: CVPR (2018)

\bibitem{zontak2011internal}
Zontak, M., Irani, M.: Internal statistics of a single natural image. In: CVPR
  2011. pp. 977--984. IEEE (2011)

\bibitem{zontak2013separating}
Zontak, M., Mosseri, I., Irani, M.: Separating signal from noise using patch
  recurrence across scales. In: Proceedings of the IEEE Conference on Computer
  Vision and Pattern Recognition. pp. 1195--1202 (2013)

\bibitem{zoran2011learning}
Zoran, D., Weiss, Y.: From learning models of natural image patches to whole
  image restoration. In: 2011 International Conference on Computer Vision. pp.
  479--486. IEEE (2011)

\end{thebibliography}
\end{document}